%% file: arxiv.tex
\documentclass[10pt]{article} % For LaTeX2e
\usepackage[preprint]{tmlr}
 
\input{math_commands.tex}

\usepackage{hyperref}
\usepackage{url}
\usepackage{graphicx}
 
\usepackage[capitalize,noabbrev,nameinlink]{cleveref}
\crefname{equation}{}{}
\crefformat{equation}{(#2#1#3)}
\Crefname{algocfline}{Algorithm}{Algorithms}
\Crefname{algocf}{line}{lines}
\Crefname{assumption}{Assumption}{Assumptions}
\crefrangeformat{assumption}{Assumptions~#3#1#4-#5#2#6}
\crefrangeformat{equation}{(#3#1#4)-(#5#2#6)}
 
\usepackage{amsmath}
\usepackage{amssymb}
\usepackage[normalem]{ulem}
\usepackage{xcolor}
\usepackage[section]{placeins} 
\usepackage{dirtytalk}
\usepackage{booktabs}
\usepackage{geometry}
\usepackage{mathtools}
\usepackage{wrapfig}
\usepackage{comment}
\usepackage{multirow}
\usepackage{subcaption}
 
\usepackage{titlesec}

\titlespacing*{\subsection}
  {0pt} % Left margin space
  {0pt} % Space above the subsection title
  {0pt} % Space below the subsection title

\newcommand{\state}{\mathbf{x}}
\newcommand{\ctrl}{\mathbf{u}}
\newcommand{\query}{\mathbf{y}}
\newcommand{\goal}{\mathbf{x_g}}
\newcommand{\operator}{\mathcal{T}}
\newcommand{\solnoperator}{\mathcal{S}}
\newcommand{\dynam}{\mathbf{g}}
\newcommand{\costparam}{{\boldsymbol{\phi}}}
\newcommand{\dynamparam}{{\boldsymbol{\psi}}}
\newcommand{\basis}{b}
\newcommand{\coeff}{c}
\newcommand{\cost}{\ell}

\title{Neural Operators for Multi-Task Control and Adaptation}
\author{\name David SeWell \email dsewell@utexas.edu \\
      \name Xingjian Li \email xingjian.li@austin.utexas.edu \\
      \name Stepan Tretiakov \email stepan@utexas.edu \\
      \name Krishna Kumar \email krishnak@utexas.edu \\
      \name David Fridovich-Keil \email dfk@utexas.edu \\
      \addr The University of Texas at Austin    }
 
\def\month{MM}  % Insert correct month for camera-ready version
\def\year{YYYY} % Insert correct year for camera-ready version
\def\openreview{\url{https://openreview.net/forum?id=XXXX}} % Insert correct link to OpenReview for camera-ready version

\begin{document}
\frenchspacing

\maketitle
 
\begin{abstract}
Neural operator methods have emerged as powerful tools for learning mappings between infinite-dimensional function spaces, yet their potential in optimal control remains largely unexplored. We focus on multi-task control problems, whose solution is a mapping from task description (e.g., cost or dynamics functions) to optimal control law (e.g., feedback policy).
 We approximate these solution operators using a permutation-invariant neural operator architecture. Across a range of parametric optimal control environments and a locomotion benchmark, a single operator trained via behavioral cloning accurately approximates the solution operator and generalizes to unseen tasks, out-of-distribution settings, and varying amounts of task observations. We further show that the branch--trunk structure of our neural operator architecture enables efficient and flexible adaptation to new tasks. We develop structured adaptation strategies ranging from lightweight updates to full-network fine-tuning, achieving strong performance across different data and compute settings. Finally, we introduce meta-trained operator variants that optimize the initialization for few-shot adaptation. These methods enable rapid task adaptation with limited data and consistently outperform a popular meta-learning baseline. Together, our results demonstrate that neural operators provide a unified and efficient framework for multi-task control and adaptation.

\end{abstract}
 
\section{Introduction}
 
In many control applications, one must solve not a single optimization problem but a family of related ones, such as navigating to different goal locations, tracking reference trajectories under varying vehicle parameters, or planning paths through environments with different obstacle and terrain configurations. Each such variation defines a distinct task, specified by the choice of dynamics, cost function, or constraints that characterize the control problem. The fundamental challenge is that even modest changes in these specifications can induce substantially different optimal policies, so a successful multi-task approach must capture how changes in task specifications translate into changes in the desired policy. This relationship is often highly nonlinear, making naive parameter-sharing or task-conditioning strategies insufficient in many practical scenarios. Furthermore, when the amount of available task data varies, some tasks may have many expert demonstrations while others have few. The learned model must be robust to this heterogeneity. This structure suggests that multi-task control is naturally formulated as a mapping between function spaces: from task-defining functions to optimal policies. This perspective motivates our use of neural operators, which are designed to approximate mappings between function spaces and are thus a natural model class for multi-task control.
 
From this perspective, common approaches to multi-task control can be viewed as approximations to the underlying structure of the problem: a mapping between infinite-dimensional function spaces. Methods that learn a separate policy per task \citep{teh2017distral, haldar2023polytask} ignore the shared structure across tasks and scale poorly. Task-embedding approaches represent tasks as finite-dimensional vectors, but their performance can be sensitive to architectural choices and variability in the number of data points for the task-defining function. Meta-learning methods such as Model-Agnostic Meta-Learning MAML \citep{finn2017model} learn a parameter initialization that enables rapid adaptation to new tasks with few gradient steps. Unlike neural operators, however, MAML does not explicitly model the mapping between function spaces. In contrast, neural operators are designed to learn mappings between function spaces directly, enabling zero-shot predictions at inference time. This makes them particularly well suited for settings where new policies must be produced quickly across many tasks at deployment. 

Neural operators have been successfully applied across a wide
range of scientific computing problems, including weather
forecasting, fluid dynamics, and materials
science~\citep{kovachki2023neural, nghiem2023physics,
lu2021learning, pathak2022fourcastnet, tretiakov2025setonet,
li2020fourier}, supported by universal approximation guarantees
for continuous operators between Banach
spaces~\citep{chen1995universal}. A useful property of several neural operator (NO) architectures is their ability to represent
complex, nonlinear mappings between function spaces while
remaining invariant to both the cardinality and ordering of
samples in the input. Despite this natural fit, their application
to control problems remains largely unexplored. In this work, we
investigate the SetONet architecture~\citep{tretiakov2025setonet},
built on DeepONet~\citep{lu2021learning}, as a model for
multi-task control. We train an operator via behavioral cloning
and show that it accurately approximates the mapping from
task-defining functions to optimal policies.

In practice, the pretrained operator may encounter tasks that
differ from those seen during training. To handle this, we develop
several adaptation strategies that exploit the branch--trunk
decomposition of SetONet, ranging from last-layer updates to
full-network fine-tuning. We further propose two meta-training
variants, SetONet-Meta and SetONet-Meta-Full,
which optimize the operator initialization for rapid few-shot
adaptation via a bi-level objective inspired by
MAML~\citep{finn2017model}. We evaluate all methods on four
parametric optimal control environments and a locomotion task from
the iMuJoCo benchmark (HalfCheetah-v3)~\citep{patacchiola2023comparing}.
Our main contributions are:
 
\begin{figure}
    \centering
    \includegraphics[width=0.75\linewidth]{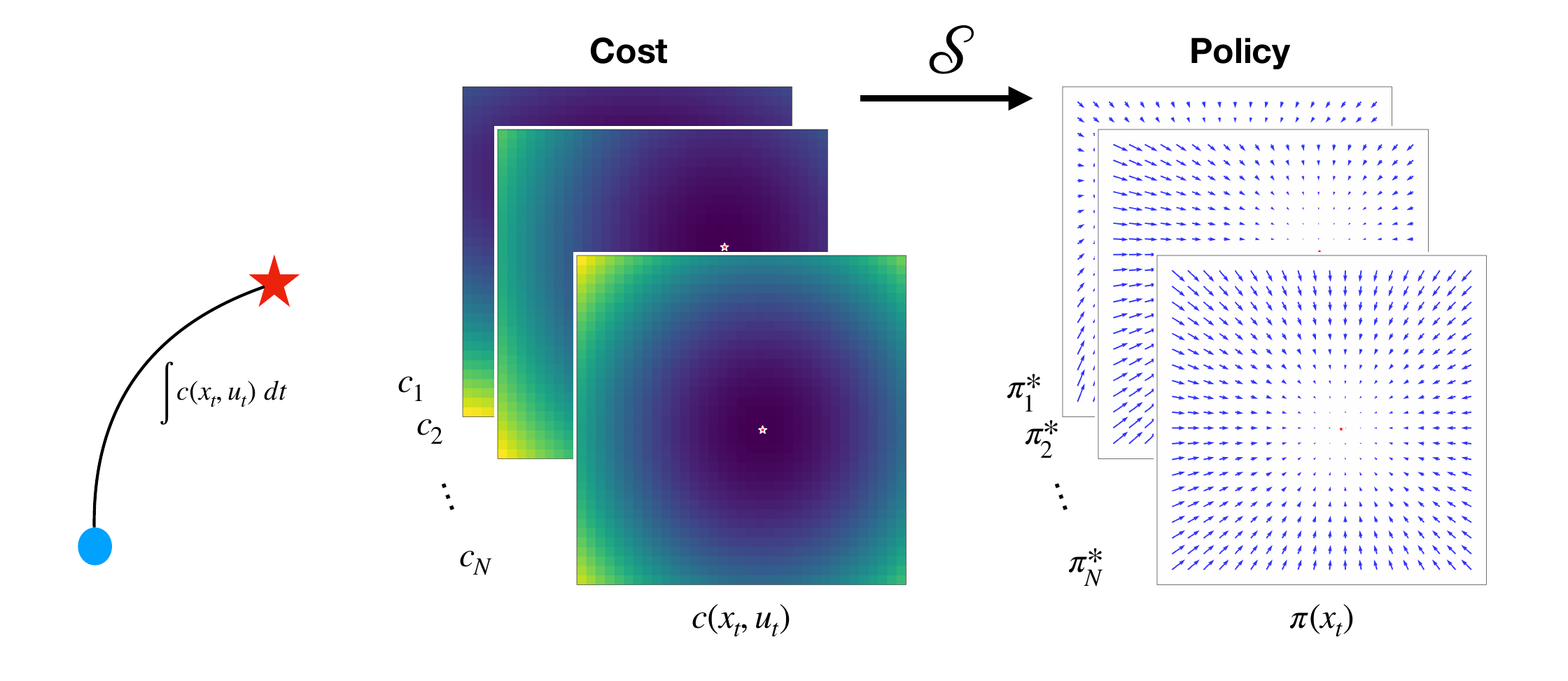}
    \caption{A point-to-point multi-task control problem. \textbf{Left:} A single point-to-point task. \textbf{Middle:} A series of tasks represented by their cost functions $c_i$. \textbf{Right:} The corresponding optimal policies $\pi^*_i$ where the small arrows correspond to the control outputs at different x,y positions. This is naturally modeled as a mapping between function spaces.}
    \label{fig:motivating_lqr}
\end{figure}

\begin{itemize}
    \item \textbf{Neural operators for multi-task control.} We establish
    neural operators as an effective model for multi-task control, a
    setting that has been relatively underexplored in the neural operator
    literature. We demonstrate through a series of experiments on environments of varying complexity that NOs can accurately approximate the
    solution operator over a distribution of tasks. Moreover, because the architecture operates on variable-size, unordered task data, the learned operator generalizes across amounts of data not seen during training. 
 
    \item \textbf{Structured adaptation for new tasks.} While the pretrained operator performs well on tasks near the training distribution, accuracy degrades on out-of-distribution tasks or when training data is limited.
    We show that the neural operator architecture enables a spectrum of
    efficient adaptation strategies, from lightweight last-layer updates
    to full-network fine-tuning. Partial fine-tuning achieves accuracy
    comparable to full-network fine-tuning at a fraction of the cost,
    and cost-based fine-tuning allows adaptation without any explicit expert demonstrations. 
 
    \item \textbf{Meta-trained operators for few-shot adaptation.} We
    propose two novel meta-training variants that optimize the operator
    initialization for rapid adaptation. SetONet-Meta restricts
    the inner loop to a subset of model parameters, providing
    data-efficient updates that are especially effective when pretraining
    data is limited or adapting to tasks that are close to the training distribution. SetONet-Meta-Full is a model that adapts the full set of model parameters, enabling rapid generalization to tasks outside the training distribution. Each of the meta-training models has trade-offs depending on initial training data, online compute limitations, and whether the downstream task is OOD. Both variants consistently outperform the MAML baseline across all environments.
\end{itemize}
 
\section{Related Work}
 
\subsection{Data-Driven Optimal Control}
 
Optimal control plays a foundational role in modeling and decision-making for complex dynamical systems, with widespread applications in robotics, aerospace, and process engineering~\citep{betts2010practical,bertsekas2012dynamic,rawlings2020model}.
Classical approaches such as LQR and MPC provide strong theoretical guarantees, but rely on repeatedly solving computationally intensive optimization problems online, which can limit their scalability to high-dimensional and highly nonlinear systems.
These challenges have motivated learning-based and data-driven approaches that aim to approximate optimal control policies through offline learning, effectively amortizing the cost of repeated online optimization. A prominent strategy is to leverage the demonstrations generated by expert controllers and to recast control as a supervised learning problem~\citep{hertneck2018learning,karg2020efficient,chen2018approximating}.
This paradigm enables faster inference at runtime than classical methods and is often more suitable for real-time control tasks. Leveraging known dynamics and expert demonstrations, training the policy model can often be done more efficiently and accurately compared to model-free reinforcement learning~\citep{reddy2019sqil}.

\subsection{Multi-Task Control}\label{sec:related_work_multi_task}
Learning a controller that can adapt to different tasks remains a challenge. When the total number of tasks is limited, a common approach, particularly in reinforcement learning, is to learn a separate policy for each task. This is often followed by a \emph{distillation} step, in which information shared across tasks is consolidated into a single policy, as proposed in~\citep{teh2017distral}. Variants of this distillation paradigm have been shown to improve performance by explicitly conditioning the final policy model with its task label~\citep{haldar2023polytask}. Although effective when the number of tasks is small and sufficient training data is available, such approaches struggle to scale: the computational cost of training and storing separate policies becomes prohibitively high for complex problems. Neural operators sidestep these scalability issues by learning a single mapping from task descriptions to policies, avoiding the need to train or store separate models as the number of tasks grows. 
 
Given such limitations, it is often preferable to learn a central policy that can perform well across a family of tasks~\citep{ammar2014online, deisenroth2014multi}.
For parametric optimal control problems in which each task is uniquely defined by a set of problem-specific parameters, \cite{drgovna2024learning} propose augmenting the policy’s state inputs with vectors of explicit task parameterizations. Although straightforward to implement, this assumption can be difficult to satisfy in practice, particularly in reinforcement learning settings where accurate task models are not always accessible.
A common alternative is to introduce learned task representations~\citep{sodhani2021multi, humplik2019meta, rakelly2019efficient, marza2024task, lan2019meta, hansen2023td}. These approaches typically learn a task embedding separately from the policy, mapping task-relevant information to a fixed-dimensional vector.
These learned representations can then be used in the policy training as conditioning explicitly or implicitly.
Although effective in many settings, this formulation introduces a representational bottleneck: task descriptions must be compressed into a fixed-length vector, making performance sensitive to observation resolution, ordering, and task coverage. Moreover, the assumptions required for convergence guarantees and other theoretical results are often quite restrictive \citep{tutunov2018distributed}. 

More recently, set-based and attention-based architectures have been
explored to handle variable-size task inputs in reinforcement
learning. \citet{mern2020exchangeable} propose an attention-based
input representation that is invariant to the ordering and number of
objects in the observation, improving sample efficiency in
environments with exchangeable entities. \citet{zhou2022policy}
formalize this setting through entity-factored Markov decision processes (MDPs) and show that
Deep Set and Self-Attention policy architectures enable compositional
generalization to varying numbers of entities at test time. These
approaches share some properties with the SetONet model used in our work; 
namely, a permutation-invariant encoder that accommodates
variable-size inputs. However, they operate
at the level of state representations and policy conditioning rather than learning a mapping between function
spaces. The operator methods used in our work provide a principled framework for learning the infinite-dimensional mappings between tasks and control policies. 
 
Lastly, meta-learning provides an alternative paradigm for
multi-task control, achieving task adaptation not through
architecture design but through a training objective that
explicitly optimizes for rapid adaptation from few examples. Methods such as MAML~\citep{finn2017model} and its
variants~\citep{collins2020task, barman2024exploring} aim to
learn a policy initialization from which task-specific policies
can be obtained with limited data and a small number of gradient
steps, without requiring explicit task representations. While conceptually appealing, these approaches incur additional computation overhead at online deployment, limiting their use in real-time control settings, where fast and reliable adaptation is critical. In contrast to these approaches, neural operators offer a principled approach that avoids both the representational bottleneck of fixed-dimensional task embeddings and the computational overhead of gradient-based adaptation at deployment. In \cref{sec:exp:meta_op}, we empirically demonstrate these advantages by comparing NO based policies against a MAML baseline across several adaptation scenarios, including zero-shot generalization and rapid online fine-tuning. 
 
\subsection{Neural Operators}

Neural operators (NOs) have emerged as a principled approach for learning mappings between infinite-dimensional function spaces, offering strong theoretical support~\citep{kovachki2023neural,kovachki2021universal,lu2021learning} and robust empirical performance across a wide range of differential equations and broader scientific computing problems~\citep{azizzadenesheli2024neural,choi2024applications,rashid2022learning,li2025zero}. 
By formulating task adaptation directly as an operator learning problem, neural operators naturally align with multi-task optimal control settings and enable efficient task-dependent adaptation with minimal online computational overhead.
Despite this potential, their application to control problems remains relatively underexplored.
Recent work has shown that solution operators can be learned in an unsupervised manner for mean-field games ~\citep{huang2024unsupervised}, with further error analysis and demonstrations of viability for certain open-loop control problems provided in~\cite{xu2025self}.
However, much of the existing literature still focuses on single-task control formulations~\citep{bhan2023neural}, highlighting the need for broader investigation of neural operators as policy learners for general parametric control problems.
Additionally, while comparisons and integrations of MAML-style meta-learning and neural operators have been explored in simulation and design contexts~\citep{wang2024meta}, their relative advantages and trade-offs in control applications remain poorly understood. Our work aims to address these gaps.

\section{Preliminaries}
 
In this section, we formulate the class of parametric optimal control problems that defines the multi-task control setting (\cref{sec:pOCP}), discuss imitation learning: a method for approximating optimal policies given expert demonstrations (\cref{sec:prelim_il}) and introduce the neural operator architectures DeepONet and SetONet (\cref{sec:prelim_neural_operators}).
 
\subsection{Parametric Optimal Control Problems}\label{sec:pOCP}
 
Consider a dynamical system with state $\state_t \in X \subset \mathbb{R}^{d_\state}$ and control input $\ctrl_t \in U \subset \mathbb{R}^{d_\ctrl}$ at discrete times $t \in [T] := \{0, 1, \dots, T\}$. 
Let $\Phi \subset \mathbb{R}^{d_\costparam}$ and $\Psi \subset \mathbb{R}^{d_\dynamparam}$ denote the sets of cost and dynamics parameters. Each task is specified by a pair $(\costparam,\dynamparam) \in \Phi \times \Psi$, where $\costparam$ determines the stage and terminal costs $(\ell,\ell_T)$ and $\dynamparam$ determines the dynamics $\dynam$. We seek a feedback policy
\mbox{$\pi \colon X \times [T - 1] \times \Phi \times \Psi \to U$}
that solves the following discrete-time parametric optimal control problem (pOCP):
\begin{equation}
\pi^*(\cdot \;;\costparam,\dynamparam)
\in
\arg\inf_{\pi}
\;
\mathbb{E}_{\state_0 \sim P_{\state_0}}
\left[
\sum_{t=0}^{T-1}
\ell\big(\state_t, \pi(\state_t,t;\costparam,\dynamparam), t; \costparam\big)
+
\ell_T(\state_T; \costparam)
\right],
\label{eq:pocp_objective}
\end{equation}
where $P_{\state_0}$ denotes some known probability distribution over the initial state $\state_0$,  the state trajectory evolves according to the closed-loop dynamics
\begin{equation}
\begin{aligned}
\state_{t+1}
&=
\dynam\big(\state_t, \pi(\state_t,t;\costparam,\dynamparam), \dynamparam\big)
\quad \forall \; t \in [T - 1], \\
\end{aligned}
\label{eq:pocp_dynamics}
\end{equation}
and the control input at time $t$ is generated by the  feedback policy $\pi$, i.e.
\begin{equation}
\begin{aligned}
\ctrl_t
&=
\pi(\state_t, t; \costparam, \dynamparam) \quad \forall \; t \in [T - 1].
\end{aligned}
\label{eq:pocp_policy}
\end{equation}

For any fixed task $(\costparam, \dynamparam)$, and initial fixed
state $\state_0$, the objective~\cref{eq:pocp_objective} together
with~\cref{eq:pocp_dynamics,eq:pocp_policy} define a standard
finite-horizon optimal control problem, which can be solved via
classical approaches, e.g.\ shooting or
collocation~\citep{jacobson1970ddp, hargraves1987direct}. Although each task is generated by a finite-dimensional parameter, we assume that there is no direct access to $(\costparam, \dynamparam)$ and instead observe only pointwise evaluations of the induced cost or dynamics functions. Across the family of tasks, the resulting
solutions define a mapping $\solnoperator$ from the task-defining functions
parameterized by $(\costparam, \dynamparam)$ to the corresponding
optimal feedback law $\pi^* \in \Pi = \{\pi^*(\cdot \;; \costparam, \dynamparam): \; (\costparam, \dynamparam) \in \Phi \times \Psi\}$; this is precisely the object we will aim to approximate with neural operators. To isolate the effects of cost and dynamics on the solution operator, we vary only one set of parameters at a time, with $\costparam \sim P_\Phi$ and $\dynamparam \sim P_\Psi$ sampled independently. We therefore focus on approximating the following two solution operators:

\begin{equation} \label{eq:soln_operator}
    \solnoperator_\dynamparam : \mathcal{F} \to \Pi, \qquad
    \solnoperator_\costparam : \mathcal{G} \to \Pi,
\end{equation}
% Spell out what F and G are explicitly here
For $\solnoperator_\dynamparam$, we assume fixed dynamics parameters $\dynamparam$, and for $\solnoperator_\costparam$, fixed cost parameters $\costparam$. The cost function space is denoted $\mathcal{F} = \{\,(\ell(\cdot\,;\costparam),\, \ell_T(\cdot\,;\costparam)) : \costparam \in \Phi\,\}$ and the dynamics function space $\mathcal{G} = \{\,\dynam(\cdot\, ;\dynamparam): \dynamparam \in \Psi \,\}$. Note that even though the input function space for $\solnoperator_\dynamparam$ does not depend on the dynamics of the system, $\Pi$ always does. Here, we assume that $\mathcal{F}$ and $\mathcal{G}$ are Banach spaces. Concretely, $\solnoperator_\dynamparam$ maps from $\mathcal{F}$ onto $\{\pi^*(\cdot\;;\costparam, \dynamparam) : \costparam \in \Phi\} \subset \Pi$, and $\solnoperator_\costparam$ maps from $\mathcal{G}$ onto $\{\pi^*(\cdot\;;\costparam, \dynamparam) : \dynamparam \in \Psi\} \subset \Pi$. The setting in which the input and output function spaces are accessible only through finite collections of pointwise samples is precisely the regime addressed by neural operator architectures, which we introduce in \cref{sec:prelim_neural_operators}.

\subsection{Imitation Learning}
\label{sec:prelim_il}
 
Given access to task functions $\ell(\cdot \; ;\costparam), \ell_T(\cdot \; ; \costparam)$ and $\dynam(\cdot \; ;\dynamparam)$ and an expert solver for the parametric optimal control problem~\cref{eq:pocp_objective}, we can generate demonstrations
consisting of $m$ state--action pairs 
$\{(\state_j, t_j, \pi^*(\state_j, t_j; \costparam, \dynamparam))\}_{j=1}^m$ where
$\ctrl_j = \pi^*(\state_j, t_j; \costparam, \dynamparam)$ for a
particular task $(\costparam, \dynamparam)$. Here we use $\pi^*$ to denote the optimal policy to the problem. \emph{Behavioral cloning}
(BC) \citep{pomerleau1988alvinn}, is a widespread technique for training a parametric policy $\hat{\pi}_\theta$, by minimizing the following loss:
\begin{equation}
  \mathcal{L}_{\mathrm{BC}}(\theta)
  = \frac{1}{m}\sum_{j=1}^{m}
    \bigl\|\hat{\pi}_\theta(\state_j, t_j, \costparam, \dynamparam) - \pi^*(\state_j, t_j;
    \costparam, \dynamparam)\bigr\|^2.
  \label{eq:bc_loss}
\end{equation}
In the single-task setting, this results in a policy that mimics the expert for one fixed choice of $(\costparam, \dynamparam)$. This form of imitation learning is appealing because it reduces the problem of learning a policy to a supervised learning problem.
 
In the \emph{multitask} setting, demonstrations are collected across a
family of tasks; for example,
$\costparam_i \sim P_\Phi$ for $i = 1,\dots, N$ with $\dynamparam$
held fixed. The goal is then to learn a single model that, given
the information identifying the current task, produces a task-specific
policy. A na\"ive approach is to train a separate policy per task, but
this scales poorly with the number of tasks and does not exploit any shared structure in the family of tasks. Alternatively, a single policy can be conditional on a task representation, but this introduces challenges discussed in  \cref{sec:related_work_multi_task}.
The operator learning perspective developed in
\cref{sec:bc_operator} offers a principled
alternative: rather than conditioning on a finite-dimensional task embedding, the model receives pointwise
evaluations of the task-defining function and learns a mapping to a space of expert policies. This multi-task behavior cloning approach retains the simplicity of \cref{eq:bc_loss} while still being able to learn the shared structure across tasks. 
 
\subsection{Neural Operators}\label{sec:prelim_neural_operators}
 
In this work, we will approximate the solution operators
of~\cref{eq:soln_operator} primarily using
SetONet~\citep{tretiakov2025setonet}, a neural operator architecture
that builds on DeepONet~\citep{lu2021learning}. DeepONet is grounded
in the universal approximation theorem for
operators~\citep{chen1995universal}, which establishes that
continuous operators between Banach spaces can be approximated to
arbitrary accuracy by a neural network comprising two sub-networks
called the \emph{branch network} and \emph{trunk network}. SetONet
inherits this theoretical foundation and additionally introduces a
permutation-invariant set encoder in the branch network, making it
naturally compatible with variable-sized, unordered input
data, properties that are particularly advantageous in control
settings where the number and arrangement of task observations may
vary. Other popular operator learning approaches such as
Fourier Neural Operators (FNO)~\citep{li2020fourier} and Graph Kernel
Networks~\citep{li2020neural} impose spectral or locality-based
inductive biases that, while effective for many PDE problems, are
less clearly motivated in parametric optimal control. 
 
Here we briefly describe the SetONet setup and architecture.
Consider the solution operator
$\solnoperator_\dynamparam : \mathcal{F} \to \Pi$ (the
dynamics-varying case is analogous). An input function
$\ell_i \in \mathcal{F}$ is observed only by pointwise evaluations
at a finite number of \emph{context points}
$\state \in \mathbb{R}^{d_x}$, and the output expert policy
$\pi^* \in \Pi$ is predicted at a finite number of \emph{query
locations} $\query \in \mathbb{R}^{d_y}$. The input and output functions may require multiple inputs as context ($\state_t, \ctrl_t, t$), which we omit for brevity in this section. SetONet comprises a
\emph{trunk network}, which learns a collection of $p$ basis
functions $\{\basis_1, \basis_2, \dots, \basis_p\}$ in the policy
space $\Pi$, and a \emph{branch network} that maps a variable-sized
set of input function evaluations to the associated coefficients
$\{\coeff_1, \coeff_2, \dots, \coeff_p\}$ via a
permutation-invariant set encoder. The basis functions $\basis_k$
may be vector-valued (e.g., $\basis_k \in \mathbb{R}^{d_u}$ for
control outputs) and the coefficients
$\coeff_k \in \mathbb{R}$. The operator is approximated as:
\begin{equation}
\solnoperator_\dynamparam(\ell_i)(\query) \approx
\operator_\theta(\ell_i)(\query) = \sum_{k=1}^{p}
\coeff_k(\ell_i)\,\basis_k(\query),
\end{equation}
 
where $\theta$ are the learned parameters of both the branch
network ($\theta_{branch}$) and the trunk network
($\theta_{trunk}$). Labels come in the form of samples from
the input and output function spaces $\ell_i \in \mathcal{F}$ and
$\pi^*_i \in \Pi$ evaluated at a finite number of sensor locations
and query locations. Concretely, for each training instance $i$ we
assume access to (i) a \emph{context set} of samples of an input
function $\ell_i \in \mathcal{F}$ at $m_i$ different sensor
locations $\{\state_{ij}\}_{j=1}^{m_i}$,
 
 \begin{equation}
 C_i \;=\; \{(\state_{ij},\, \ell_i(\state_{ij}))\}_{j=1}^{m_i},
\end{equation}
 and (ii) targets consisting of evaluations of the output policy
$\pi^*_i = \solnoperator_\dynamparam(\ell_i) \in \Pi$ at $n_i$
different query locations $\{\query_{ik}\}_{k=1}^{n_i}$, e.g.,
\begin{equation}
 \{(\query_{ik},\, \pi^*_i(\query_{ik}))\}_{k=1}^{n_i}.
\end{equation}

The neural operator $\operator_{\theta}$ is trained by minimizing
the following empirical MSE over $K$ pairs of input-output
functions $\{(\ell_i, \pi^*_i)\}_{i=1}^K$:
 \begin{equation}
  \mathcal{L}(\theta) =
  \frac{1}{K}\sum_{i=1}^K\frac{1}{n_i}\sum_{k=1}^{n_i}
  \|\operator_{\theta}(\ell_i)(\query_{ik}) -
  \pi^*_i(\query_{ik})\|_2^2
  \label{eq:emploss}
 \end{equation}
 
\begin{figure}
    \centering
    \includegraphics[width=0.85\linewidth]{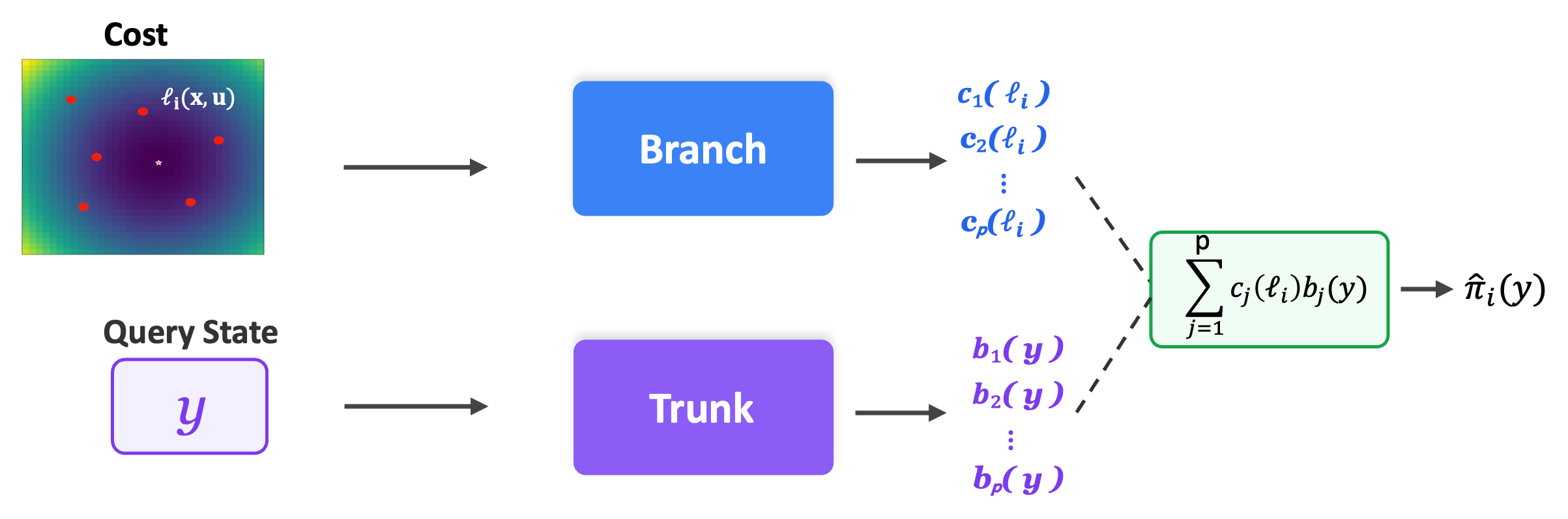}
    \caption{\textbf{DeepONet/SetONet architecture:} Here we show the mapping $\operator_\theta[\ell_i] \rightarrow \hat{\pi}_i$, with pointwise evaluations of $\ell_i$ (in red) and of $\hat{\pi}_i$ at the point $\query$. The branch network maps sensor locations $(x, u)$ of a cost function $\cost(x, u; \costparam)$ to task-dependent coefficients $\{\coeff_k(\ell)\}_{k=1}^p$. The red points indicate the pointwise samples of $\cost_i$. The trunk maps query locations $\query = (x, t)$ to a set of learned basis functions $\{\basis_k(\query)\}_{k=1}^p$. Their inner product yields the predicted control output at the query location $\query$}
    \label{fig:DeepOnet_architecture}
\end{figure}

This decomposition is illustrated in
\cref{fig:DeepOnet_architecture} for the case where $\mathcal{F}$
is the space of cost functions and the output space $\Pi$ is the
space of optimal control policies. Because the branch network uses
a set encoder, predictions are invariant to the ordering of
samples in $C_i$ and the context set size $m_i$ may vary freely
across tasks and over time. There are no restrictions on the query
locations, which can be arbitrarily chosen at training or test
time. Although the expert trajectories used in our experiments are generated by open-loop solvers, the operator learns a feedback mapping conditioned on the current
state. Along the expert trajectory, the open-loop and closed-loop
representations coincide, and the additional state dependence allows
the learned policy to generalize beyond the nominal trajectory.
\section{Methodology}
 
We now describe how the behavioral cloning objective of \cref{sec:prelim_il}
and the neural operator approach of \cref{sec:prelim_neural_operators} combine
to produce a practical method for multi-task control.
In \cref{sec:bc_operator}, we show how the SetONet architecture can be
trained to approximate the solution operators defined
in~\cref{sec:pOCP} using expert demonstrations collected
across a distribution of tasks.
In principle, the pretrained operator can predict policies for new tasks
in a single forward pass. In practice, however, approximation error,
limited training data, and distributional shift between training and
deployment tasks can degrade zero-shot predictions.
We therefore develop strategies for adapting the pretrained operator
to new tasks at deployment:
fine-tuning with expert demonstrations, fine-tuning with cost feedback
when expert demonstrations are unavailable (\cref{sec:transfer}), and a
meta-training procedure that explicitly optimizes the operator for rapid
few-shot adaptation (\cref{sec:meta_training}).
 
\label{sec:methodology}
 
\subsection{Multi-Task Behavioral Cloning via Operator Learning}
\label{sec:bc_operator}
 
We now instantiate the behavioral cloning objective
of~\cref{sec:prelim_il} within the operator learning approach
of~\cref{sec:prelim_neural_operators} to approximate the solution operators $\solnoperator_\dynamparam$ and
$\solnoperator_\costparam$ defined in~\cref{eq:soln_operator}.
 
\paragraph{Data generation.}
For concreteness, we will consider the cost-varying setting in which the
dynamics $\dynamparam$ are fixed and
$\costparam_i \sim P_\Phi$ for $i = 1, \dots, N$ (the dynamics-varying
case is analogous). For each sampled task $\costparam_i$, we obtain an expert policy
$\pi_i^{\mathrm{exp}} \approx \pi^*_i$ using a suitable optimal control solver
(see \cref{sec:exp_env} for specific solvers used) or RL algorithm
(Soft Actor Critic (SAC) for iMuJoCo). We write $\pi^{exp}_i$ to make clear that the policy we are sampling may not exactly match $\pi^*_i$. In the OCP environments,
$\pi_i^{\mathrm{exp}}$ closely approximates the optimal feedback
law; in the iMuJoCo environments, it is a learned policy that may
be suboptimal. We consider each of these solvers as producing our
expert training target $\pi_i^{\mathrm{exp}}$. For brevity, we write
$\cost_i$ for the cost function of the $i$-th task, encompassing both
stage $\cost_i(\state_t, \ctrl_t, t)$  and terminal $\cost_i(\state_t, T)$ costs 
$\big(\ell(\cdot\;;\costparam_i),\,\ell_T(\cdot\;;\costparam_i)\big)$.  
Each training instance is thus a pair of functions: a task-defining
input $\cost_i$ and the corresponding expert policy
$\pi_i^{\mathrm{exp}}$.
 
% clarify in the context set part that \ell_i depends on what t_ij is
\paragraph{Neural operators for behavioral cloning}
Instantiating the SetONet model of~\cref{sec:prelim_neural_operators},
we construct for each task $i$:
 
\begin{enumerate}
  \item A \emph{context set}
        $C_i = \{((\state_{ij}, \ctrl_{ij}, t_{ij}),\,
        \cost_i(\state_{ij}, \ctrl_{ij}, t_{ij}))\}_{j=1}^{m_i}$
        of pointwise cost evaluations at $m_i$ sensor locations,
        which the branch network uses to encode the task identity
        without requiring access to $\costparam_i$.
  \item Supervised targets
        $\{(\query_{ik},\,\pi_i^{\mathrm{exp}}(\query_{ik}))\}_{k=1}^{n_i}$
        of expert policy evaluations at $n_i$ query locations,
        where $\query_{ik} = (x_{ik}, t_{ik})$.
\end{enumerate}

\noindent
The context set structure above applies directly to the cost-varying setting, but our model accommodates other task specifications with
only a change in what the context encodes. In the dynamics-varying setting (with cost $\costparam$ fixed, $\dynamparam_i \sim P_\Psi$), the context
set consists of dynamics evaluations $C_i = \{((\state_{ij}, \ctrl_{ij}),\,
\dynam(\state_{ij}, \ctrl_{ij};\dynamparam_i))\}_{j=1}^{m_i}$, where each context point is a state--control pair and the observed value
is the resulting next state. While we focus on specific representations of the input function for the cost-varying and dynamics-varying settings described above, we note
that the choice of context encoding is not unique. Alternative
representations, whether explicit or implicit, can be used provided
they sufficiently characterize the task. Similarly, for a reference tracking scenario, an implicit representation of cost may be given by a set of waypoint locations. In all cases, the supervised
targets for the output function space remain expert policy evaluations, the branch network receives the task-defining context, and the trunk network receives states and times at which the policy is to be predicted. The complete details for each environment are given in \cref{sec:operator_fitting}.
 
We approximate the operator $ \solnoperator_\dynamparam \approx \operator_\theta$ by minimizing the empirical loss~\cref{eq:emploss} on $N$ sets of task contexts.
Then, given a new context set $C_{i'}$, $\operator_\theta$ predicts the corresponding feedback policy at
arbitrary query locations:
$\operator_\theta(\cost_{i'})(\query) \approx \pi_{i'}^{\mathrm{exp}}(\query)$ or $\operator_\theta(\dynam_{i'})(\query) \approx \pi_{i'}^{\mathrm{exp}}(\query)$ when the input function is the space of dynamics functions.
At test time, this amounts to approximating an expert policy on an
unseen task via a single forward pass, without re-solving the
underlying optimal control problem~\cref{eq:pocp_objective}.

The formulation above and in \cref{sec:pOCP} assumes that the context set consists of
pointwise evaluations of a known input function. In practice, this
assumption can be relaxed. For example, in our obstacle avoidance
experiments the context set encodes obstacle geometry rather than
evaluations of a smooth cost function, and in the iMuJoCo
environments the context set consists of state-action transition
samples rather than evaluations of a known dynamics model. In both
cases, the branch network receives a set of tuples that implicitly
characterize the task, and the architecture operates identically.

\subsection{Task-Specific Adaptation}
\label{sec:transfer}
 
In the multi-task setting of \cref{sec:bc_operator}, the operator
$\operator_\theta$ is trained on tasks sampled from distributions
$P_\Phi$ and $P_\Psi$. At deployment, we may encounter a task
that was not seen during training but is drawn from the same
distribution. The operator $\operator_\theta$ trained in
\cref{sec:bc_operator} provides a global approximation to the solution
operator across the task distribution. When training data is limited or the target task differs substantially from the training distribution, the pretrained operator alone may not provide sufficient accuracy.
In such cases, we can adapt the operator to the new task using a small
amount of task-specific data. We consider two general adaptation settings depending on what information is available for the new task.
 
\paragraph{Adaptation with expert demonstrations.}
When a small number of expert state--action pairs
$\mathcal{D}_i = \{(\query_{j}, \ctrl_{j}^*)\}_{j=1}^{n_i}$ are available for
the target task, we can refine the operator by minimizing the imitation
loss over these demonstrations. Given a new task $i$ with sensor
locations $\mathcal{C}_i$, we minimize
\begin{equation}
    \theta^*
    = \arg\min_{\theta}
      \frac{1}{n_i}
      \sum_{(\query_j, \ctrl_j^*) \in \mathcal{D}_i}
      \bigl\|\operator_\theta(f_i)(\query_j) - \ctrl_j^*\bigr\|^2,
    \label{eq:finetune}
\end{equation}
where $f_i$ is encoded via $\mathcal{C}_i$. This reduces adaptation to
a supervised fine-tuning problem: the pretrained operator provides a
warm start, and the demonstrations steer it toward the target task's
policy.
 
\paragraph{Adaptation with cost feedback.}
In settings where expert demonstrations are unavailable but the task-specific cost
function and dynamics model are known, we can bypass the behavioral
cloning objective entirely and instead fine-tune the operator by
directly minimizing the control objective over unrolled trajectories.
Given differentiable dynamics
$\state_{t+1} = \dynam(\state_t, \ctrl_t; \dynamparam)$ and cost
$\cost(\state_t, \ctrl_t, t; \costparam)$, we roll out the current
policy $\hat{\pi}_\theta = \operator_\theta(\cost)$ from a batch of
initial states $\{\state_0^{(k)}\}$ and minimize the total trajectory
cost:
\begin{equation}
    \mathcal{L}_{\text{cost}}(\theta) = \frac{1}{M} \sum_{m=1}^{M}
    \left[ \sum_{t=0}^{T-1}
    \ell\bigl(\state_t^{(m)}, \hat{\pi}_\theta(\state_t^{(m)}, t),
    t; \costparam\bigr)
    + \ell_T\bigl(\state_t^{(m)}; \costparam\bigr) \right],
    \label{eq:cost_finetune}
\end{equation}
where each trajectory is obtained by rolling out the policy from the
initial state $\state_0^{(m)}$ through the differentiable dynamics, and
$M$ is the number of initial conditions sampled. Gradients
of~\cref{eq:cost_finetune} with respect to $\theta$ are computed via
backpropagation through the entire rollout. This is particularly
appealing for tasks where the cost structure is known but expert
solutions are unavailable or expensive to obtain, as it leverages
the pretrained policy as a warm start and refines it using only cost
feedback and a dynamics model, bypassing the difficulty of training the policy from scratch.
 
\paragraph{Which parameters to adapt.}
Recall that
$\operator_\theta(\ell_i)(\query) = \sum_{k=1}^{p} c_k(\ell_i)\,b_k(\query)$,
where the trunk produces basis functions $\{b_k\}$ shared across all
tasks and the branch produces task-dependent coefficients $\{c_k\}$.
The branch--trunk decomposition admits a spectrum of fine-tuning
strategies that range from full parameter updates to partial
retraining of the policy
model~\citep{zhu2023reliable, goswami2022deep, xu2023transfer,
wu2024fine, zhang2024d2no}.
We consider three strategies, listed in decreasing order of
adaptation capacity:
 
\begin{itemize}
    \item \textbf{Full-network fine-tuning.}
    As a baseline, we update all trainable parameters
    $\theta = (\theta_{\mathrm{branch}}, \theta_{\mathrm{trunk}})$
    during adaptation. Full updates typically yield the lowest error
    but incur the highest computational cost and are most susceptible
    to overfitting when few demonstrations are available. In subsequent sections we refer to this as \textbf{SetONet-FT}. 
 
    \item \textbf{Branch-only fine-tuning.}
    If the target policy lies approximately in the span of the learned
    basis functions,
    $\pi(\cdot) \approx \operatorname{span}\{b_1, \ldots, b_p\}$,
    it suffices to retrain only the branch network while keeping the
    trunk fixed. This restricts adaptation to finding new coefficients
    that best represent the target task's policy in the existing basis,
    reducing both the number of trainable parameters and the risk of
    catastrophic forgetting. This strategy follows naturally from
    the operator fine-tuning literature~\citep{goswami2022deep, zhang2024d2no}
    and mirrors the common practice in robotics of fine-tuning a
    task-specific head on top of a frozen
    backbone~\citep{brohan2024rt, team2024octo}. We refer to this variant as \textbf{Full-Branch} in our experiments. 
 
    \item \textbf{Last-layer fine-tuning.}
    As the lightest-weight alternative, we freeze all parameters
    except the final output layer of the branch network (\textbf{Last-Branch}), the final
    output layer of the trunk network (\textbf{Last-Trunk}), or
    both (\textbf{Last-Both}) ~\citep{xu2023transfer, zhu2023reliable, wu2024fine}.
    By restricting updates to the output heads of each sub-network,
    this approach keeps the adaptation cost minimal while still
    allowing limited flexibility in both the coefficients and the
    basis functions.
\end{itemize}
 
\noindent
These three strategies offer a clear trade-off between adaptation
capacity and computational cost. We compare them empirically in
\cref{sec:adaptation}, where we find that branch-only
fine-tuning achieves accuracy comparable to full-network updates
at a fraction of the cost, suggesting that the pretrained basis
functions transfer well across tasks, also highlighting the practical feasibility of fast online adaption of NO controllers.
 
\subsection{Meta-Training for Rapid Adaptation}\label{sec:meta_training}
The approaches taken in \cref{sec:transfer} are based on a neural
operator that was pretrained with the standard behavioral cloning
objective~\cref{eq:bc_loss} and subsequently adapted to new tasks.
As shown in \cref{sec:experiments}, this pretrained operator
generalizes well across a range of tasks and adaptation strategies.
Here we investigate replacing the behavioral cloning loss with one
that is designed for fast adaptation. This can have benefits in a
number of settings, for instance when the amount of task-specific
data available at deployment is much smaller than what was used
during training. Additionally, when pretraining data itself is
limited, meta-training can compensate: as we show in
\cref{sec:exp:meta_op}, SetONet-Meta (meta update branch only) improves over the pretrained model on P2P-Cost-Small, where standard pretraining alone does not
fully capture the task distribution. To target these settings, we
modify the training pipeline using a bi-level formulation inspired
by MAML \citep{finn2017model} that trains a neural operator whose
initialization is specifically optimized for few-shot task
adaptation. This procedure is split into \textit{inner} and
\textit{outer} loops; we consider two variants that differ in which
parameters participate in the inner loop, yielding different
trade-offs between adaptation speed and representational flexibility.
A commonly cited limitation of meta-learning is that the learned
initialization may not generalize well to tasks far from the training
distribution. We show that by adapting both the branch and trunk
networks during the inner loop, SetONet-Meta-Full (meta update full network) is able to adapt
to an OOD task on the Quadrotor environment.
 
\paragraph{Bi-level training objective.}
We adopt the episodic training structure of MAML but apply it to the
neural operator architecture. Denote the full parameter set as
$\theta = (\theta_\text{branch}, \theta_\text{trunk})$. In each
training episode, we sample a batch of $B$ tasks
$\{i_1, \dots, i_B\}$ and, for each task~$i$, split the available
data into a \emph{support} set $\mathcal{D}_i^\text{tr}$ used for
inner-loop adaptation and a \emph{query} set
$\mathcal{D}_i^\text{eval}$ used for outer-loop evaluation. Both
sets contain context data (sensor location--value pairs for the
branch) and target state--control pairs (at query locations for the
trunk and loss computation).

\begin{figure}
    \centering
    \includegraphics[width=0.8\linewidth]{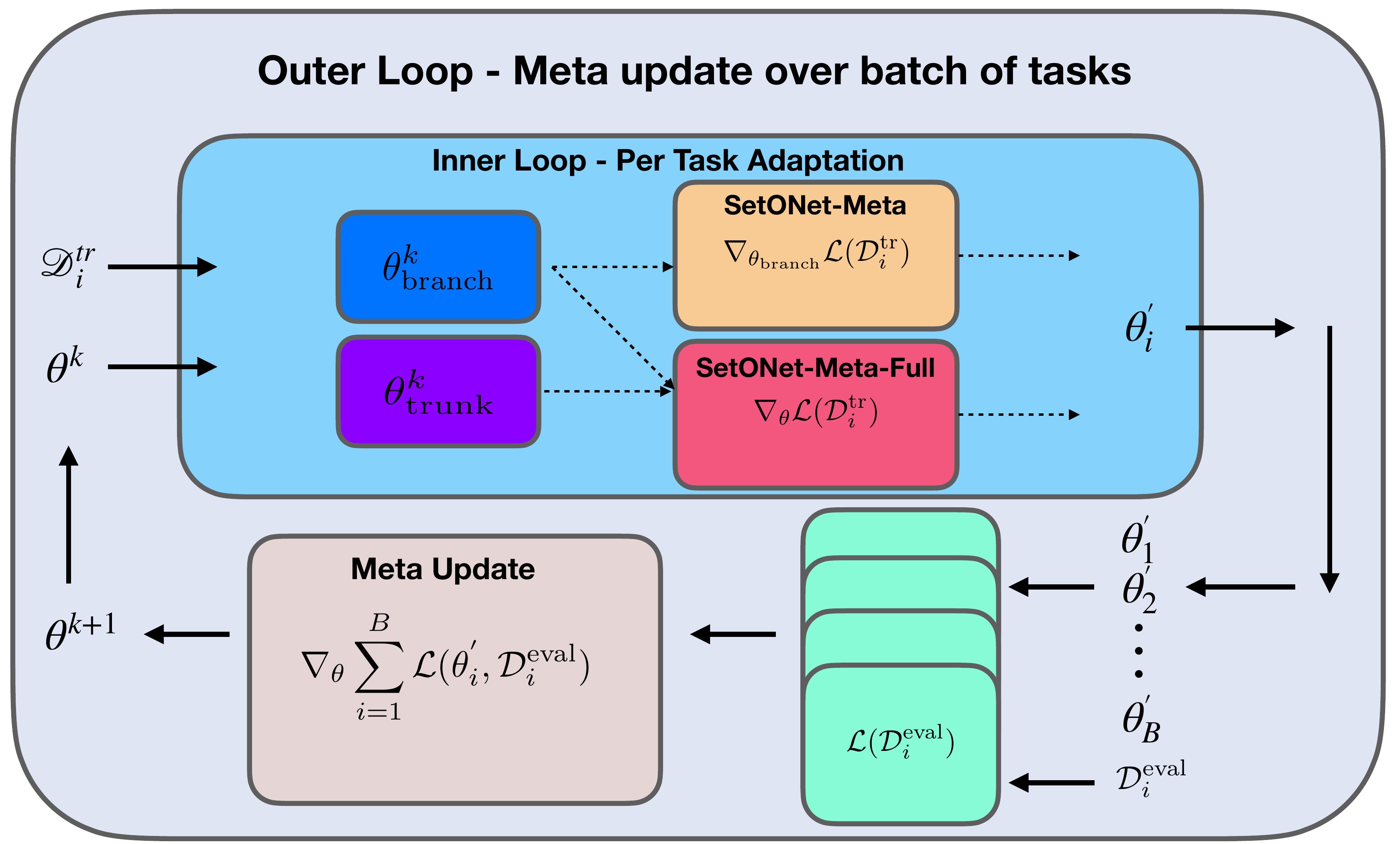}
    \caption{Overview of the meta-training procedure. The inner loop
adapts the parameters~$\theta^k$ to each task~$i$ using the support set~$\mathcal{D}_i^{\mathrm{tr}}$.
SetONet-Meta updates only the branch
coefficients~$\theta_{\mathrm{branch}}$, while
SetONet-Meta-Full updates all
parameters~$\theta$. The adapted parameters~$\theta_i'$
from a batch of~$B$ tasks are evaluated on held-out query
sets~$\mathcal{D}_i^{\mathrm{eval}}$, and the outer loop
computes the meta-gradient to update the shared initialization~$\theta^{k+1}$.}
    \label{fig:meta_diagram}
\end{figure}

The inner loop performs a single gradient descent step on the
support loss. We consider two choices for the scope of this update:
 
\begin{itemize}
    \item \textbf{SetONet-Meta} (branch-only inner loop). Only the
    branch parameters $\theta_\text{branch,i}$ for task $i$ are updated in the inner loop while the trunk
    remains frozen:
    \begin{equation}
        \theta_\text{branch,i}' = \theta_\text{branch}
        - \alpha \nabla_{\theta_\text{branch}}
        \mathcal{L}_{\mathcal{D}_i^\text{tr}}
        (\theta_\text{branch}, \theta_\text{trunk}).
        \label{eq:inner_branch}
    \end{equation}
    This preserves the learned basis functions $\{b_k\}$ and
    restricts adaptation to finding new coefficients $\{c_k\}$ that
    best represent the target task's policy. The trunk parameters
    participate only in the outer loop, where they receive gradients
    that account for how well the basis functions support adaptation
    across the full task distribution. This strategy leverages the
    branch--trunk decomposition of the operator: the trunk learns
    basis functions that are optimized to be \emph{reused} across
    tasks, while the branch learns an initialization from which a
    single gradient step yields a good task-specific policy.
 
    \item \textbf{SetONet-Meta-Full} (full-network inner loop). Both
    branch and trunk parameters $\theta_i$ for task $i$ are updated in the inner loop:
    \begin{equation}
        \theta'_i = \theta
        - \alpha \nabla_{\theta}
        \mathcal{L}_{\mathcal{D}_i^\text{tr}}(\theta),
        \label{eq:inner_full}
    \end{equation}
    This allows the operator to adapt both the coefficients
    \emph{and} the basis functions to a new task in a single step.
    The outer loop optimizes a shared initialization from which
    full-network adaptation is effective, analogous to standard
    MAML. This variant has strictly greater adaptation capacity
    than the branch-only strategy, since it can reshape the basis
    functions for each task, but it updates more parameters per
    step and may be more susceptible to overfitting when few
    demonstrations are available.
\end{itemize}
 
In both cases, the outer loop updates \emph{all}
parameters, both $\theta_\text{branch}$ and
$\theta_\text{trunk}$, by minimizing the post-adaptation loss on
the query sets, averaged across the task batch: 
\begin{equation}
    \theta \leftarrow \theta
    - \beta \frac{1}{B} \nabla_\theta \sum_{i=1}^{B}
    \mathcal{L}_{\mathcal{D}_i^\text{eval}}
    (\theta_\text{branch,i}', \theta_\text{trunk,i}'),
    \label{eq:outer}
\end{equation}
where $\beta$ is the outer learning rate. For SetONet-Meta,
$\theta_\text{trunk,i}' = \theta_\text{trunk}$ (unchanged by the
inner loop), so the outer gradient with respect to
$\theta_\text{trunk}$ flows only through the query loss evaluated
at the adapted branch. For SetONet-Meta-Full, both components pass
through the inner update via second-order differentiation. This architecture 
is shown in \cref{fig:meta_diagram} where we show the inner and outer training loops, as well as how the two proposed architectures fit in. 
 
\paragraph{Relationship to standard training and MAML.}
Both meta-training variants occupy a middle ground between the
standard pre-training pipeline of \cref{sec:prelim_neural_operators} and
classical MAML. Like MAML, they use bi-level optimization to learn
representations that are explicitly designed for rapid adaptation.
SetONet-Meta-Full is closest to standard MAML in spirit, differing
primarily in the architecture (a neural operator rather than a
monolithic policy network); it learns a full-network initialization
from which a single gradient step yields a good task-specific model.
SetONet-Meta, by contrast, exploits the branch--trunk decomposition
unique to neural operators: the trunk learns shared basis functions
that remain fixed during adaptation, and the branch learns an
initialization from which task-specific coefficients can be rapidly
recovered. This restricted inner loop reduces the number of
trainable parameters at adaptation time and lowers the risk of
overfitting, at the cost of limiting representational flexibility
to the span of the pretrained basis. 
 
\section{Experiments}
\label{sec:experiments}
 
We evaluate the neural operator model developed in
\cref{sec:methodology} on four optimal control environments of
increasing complexity and a higher-dimensional locomotion environment
from the iMuJoCo benchmark (\cref{sec:exp_env}),
where expert policies are trained via reinforcement learning. Our evaluation is
organized into two parts. First, in \cref{sec:operator_fitting}, we
demonstrate that a neural operator can accurately approximate the
solution operator across the task distribution and establish
baseline accuracy across all environments
(\cref{tab:adaptation_results}). We additionally examine how
prediction accuracy varies with the number and ordering of context
points provided at test time, including sizes not seen during
training (i.e., \emph{task
resolution invariance}). Second, in
\cref{sec:adaptation}, we evaluate the adaptation strategies of
\cref{sec:transfer,sec:meta_training}: fine-tuning with expert
demonstrations, a per-task comparison against MAML
(\cref{fig:transfer_summary}), adaptation using cost feedback on
environments where expert data is unavailable or expensive to
obtain, and a detailed study of the meta-trained operator on
HalfCheetah-v3 that examines how adaptation performance scales
with the number of demonstrations and gradient steps
(\cref{fig:halfcheetah_grid}).
 
\subsection{Environments}\label{sec:exp_env}
 
We consider four optimal control environments and a
reinforcement learning environment from the iMuJoCo benchmark,
summarized in \cref{tab:environments} and further described below.
 
The \textbf{Point-to-Point Cost (P2P-Cost)} environment is a 2D point mass with linear dynamics and quadratic running and terminal costs, where each task corresponds to a different goal state $\goal$ and the expert is a linear quadratic regulator (LQR) controller. The context set consists of state--control--cost tuples $\mathcal{C} = \{((x, u, t),\, \ell(x, u, t; \costparam))\}$ that encode the task-defining cost function. The stage and terminal costs are given by
\begin{equation}
    \ell(\state_t, \ctrl_t;\costparam) 
    = (\state_t - \state_g)^\top Q\,(\state_t - \state_g) 
      + \ctrl_t^\top R\,\ctrl_t, 
    \qquad
    \ell_T(\state_T;\costparam) 
    = (\state_T - \state_g)^\top Q_T\,(\state_T - \state_g),
    \label{eq:lqr_cost}
\end{equation}
where $Q, Q_T \succeq 0$ and $R \succ 0$ are fixed weight matrices
shared across all tasks and the task parameter
$\costparam = \state_g$ specifies the goal location.
The linear dynamics are given by
\begin{equation}
\state_{t+1} = A\state_t + B\ctrl_t, \quad \text{where} \ A = \begin{bmatrix}
    1 & 0 & \Delta t & 0 \\
    0 & 1 & 0 & \Delta t \\
    0 & 0 & 1 &        0 \\
    0 & 0 & 0 &        1 \\
\end{bmatrix}, \ B = \begin{bmatrix}
    0 & 0 \\
    0 & 0 \\
    \Delta t &        0 \\
    0 & \Delta t         \\
\end{bmatrix},
\end{equation}
where $\Delta t > 0$ is a fixed constant indicating an Euler time discretization step size. This is a standard LQR problem and a direct instance of the
parametric optimal control formulation~\cref{eq:pocp_objective}. In the experiments that follow, we include a P2P-Cost-Small that allows us to test how different models perform when there is less data to train.

\paragraph{Point-to-Point Dynamics (P2P-Dynamics)}\label{def:p2p_dynamics}
This environment uses the same point-mass state and control spaces
as P2P-Cost but introduces parameter-dependent
nonlinearities. The state is $\state = (p_x, p_y, v_x, v_y)$ and
the control $\ctrl = (a_x, a_y)$ is a commanded acceleration.
The dynamics, discretized with an Euler step of size $\Delta t$, are
\begin{equation}
    \state_{t+1} = \state_t + \Delta t
    \begin{bmatrix}
        v_{x,t} \\ v_{y,t} \\
        \text{clip}(\mu \, a_{x,t},\; -a^{\max},\; a^{\max}) \\
        \text{clip}(\mu \, a_{y,t},\; -a^{\max},\; a^{\max})
    \end{bmatrix},
    \label{eq:p2p_dyn_dynamics}
\end{equation}
where $\text{clip}(x, a, b) = \max(a, \min(x, b))$. The
control gain $\mu \in (0, 1]$ attenuates the commanded
acceleration, and per-axis acceleration saturation enforces
$\|\dot{v}\|_\infty \leq a^{\max}$. After each step,
velocities are additionally clipped to enforce
$\|v\|_\infty \leq v^{\max}$. Each task is defined by a distinct
dynamics parameterization
$\dynamparam = (\mu, v^{\max}, a^{\max})$, making this an instance of
\eqref{eq:pocp_objective} with fixed cost $\costparam$ and varying
dynamics. The context set encodes evaluations of the dynamics function:
$\mathcal{C} = \{((\state, \ctrl),\, \dynam(\state, \ctrl; \dynamparam))\}$.
Because the clipping makes the dynamics nonlinear, expert trajectories
are generated with iterative linear quadratic regulator (iLQR)
\citep{tassa2012synthesis} rather than closed-form LQR.

\paragraph{Planar Quadrotor}
The planar quadrotor has a 6D state
$\state = (y, z, \phi, \dot{y}, \dot{z}, \dot{\phi})$
and control $\ctrl = (F_z, \tau)$ consisting of the total thrust and
roll torque. The dynamics, discretized with an Euler step of
size $\Delta t$, are
\begin{equation}
\state_{t+1} = \state_t + \Delta t
\begin{bmatrix}
    \dot{y}_t \\
    \dot{z}_t \\
    \dot{\phi}_t \\
    -\tfrac{1}{m}F_{z,t} \sin\phi_t \\
    \tfrac{1}{m}F_{z,t} \cos\phi_t - g \\
    \tfrac{1}{I}\tau_t
\end{bmatrix},
\label{eq:quad_dynamics}
\end{equation}
where $I = \alpha m L^2$ is the moment of inertia.
Each task is defined by varying physical parameters
$\dynamparam \in \mathbb{R}^3 = (m, L, \alpha)$: mass, arm length,
and an inertia scaling factor. The context set encodes dynamics
evaluations
$\mathcal{C} = \{((\state, \ctrl),\, \dynam(\state, \ctrl; \dynamparam))\}$.
Expert trajectories are computed with iLQR, targeting a fixed
hover state.
 
\paragraph{Obstacle Avoidance}
This environment uses the same double-integrator dynamics as P2P-Cost 
(state $\state \in \mathbb{R}^4$, control $\ctrl \in \mathbb{R}^2$, with $A$ and $B$ as defined above),
but each task is defined by a randomly generated field of
$n_{\mathrm{obs}} \in [2, 6]$ circular obstacles, each
parameterized by its center $\mathbf{(x,y)}_i$.
The expert solves a constrained nonlinear programming problem
(NLP) via IPOPT~\citep{wachter2006implementation,andersson2019casadi},
minimizing control effort subject to hard collision avoidance
constraints.
Unlike the other environments, the task-defining function is not
directly available as a smooth cost; instead, the context set
encodes the obstacle geometry as a collection of tuples
$\mathcal{C} = \{(\state_i, \query_i, r_i)\}_{i=1}^{n_{\mathrm{obs}}}$,
which serves as a finite-dimensional proxy for the underlying
constraint structure. This is an instance of \cref{eq:pocp_objective} with fixed dynamics,
where the task variation enters through the constraint structure
rather than the cost function directly.
 
\paragraph{iMuJoCo}
To test generalization beyond the model-based optimal control
problems, we additionally evaluate on HalfCheetah-v3 from the
iMuJoCo benchmark \citep{patacchiola2023comparing}, which
provides families of MuJoCo agents whose physical
parameters: body mass, limb length, joint range, and surface
friction, are systematically varied from a default
configuration. HalfCheetah-v3 ($d_x = 17$, $d_u = 6$, 53
configurations) is a high-dimensional locomotion task where
expert policies are trained via Soft Actor-Critic
(SAC)~\citep{haarnoja2018soft} rather than computed from a
known dynamics model, and the resulting demonstrations are
collected as offline rollouts.
 
\begin{table}[t]
\caption{Summary of experimental environments. All OCP environments
use the same SetONet architecture and training procedure; only the
context and query definitions change across environments. The
iMuJoCo environment encodes tasks through state--action transition
data rather than analytical function evaluations. ``Tasks'' denotes
the number of distinct task configurations and ``Traj.'' the number
of expert trajectories per task.}
\label{tab:environments}
\centering
\footnotesize
\setlength{\tabcolsep}{4pt}
\begin{tabular}{@{}l c c r r l@{}}
    \toprule
    Environment & $d_x/d_u$ & Expert & Tasks & Traj. &
    Task parameters \\
    \midrule
    P2P-Cost & 4/2 & LQR & 500 & 100 &
    $\mathbf{x}_g \in [-10,10]^2$ \\
    P2P-Cost-Small & 4/2 & LQR & 50 & 10 &
    $\mathbf{x}_g \in [-10,10]^2$ \\
    P2P-Dynamics & 4/2 & iLQR & 100 & 100 &
    $\mu, v^{\max}, a^{\max}$ \\
    Quadrotor & 6/2 & iLQR & 100 & 20 &
    $m, L, \alpha$ \\
    Obstacle & 4/2 & NLP & 500 & 60 &
    $n_{\text{obs}}, \mathbf{(x,y)}_i$ \\
    \midrule
    HalfCheetah-v3 & 17/6 & SAC & 53 & 100 &
    mass, limb, joints, friction \\
    \bottomrule
\end{tabular}
\end{table}
 
All of our experiments are done on a single GPU (NVIDIA GeForce RTX 5070).
They all use the same SetONet architecture and training
procedure. The branch network uses a set encoder consisting of
a multilayer perceptron (MLP) $\phi$ applied to each element of the context, followed by mean-pooling aggregation
and a post-aggregation MLP $\rho$; the trunk network is a
standard MLP. The full list of hyperparameters and architecture is included in the supplementary material. 
% within the yaml config 
% files of the included codebase. 
All
models are trained using the Adam optimizer~\citep{kingma2014adam}
with a learning rate of $\eta = 10^{-3}$ to minimize the
mean-squared-error loss~\cref{eq:emploss}. During training,
the number of context points $m$ is sampled uniformly from a
predefined set at each iteration, encouraging the model to
perform well across varying context resolutions. For each
environment, the task dataset is split 80\%/20\% into training
and held-out test tasks; all reported metrics are computed on
held-out tasks not seen during training. States and context
values are normalized to zero mean and unit variance using
statistics computed from the training set. Models are
implemented in JAX using the Equinox library~\citep{kidger2021equinox}.
 
\subsection{Neural Operator Fitting and Task Resolution}
\label{sec:operator_fitting}

We first evaluate the ability of the SetONet model to approximate the
solution operator across each task distribution when trained with the
behavioral cloning objective~\cref{eq:bc_loss}. Rather than reporting
raw mean squared error (MSE), we use the relative $L^2$ error
\begin{equation}\label{eq:rel_l2}
    \text{Relative } L^2
    ~\text{error}= \frac{\|\operator_\theta(\ell_i)(\query) - \ctrl^*\|_2}
           {\|\ctrl^*\|_2},
\end{equation}
 
which normalizes by the magnitude of the expert control signal and
provides an interpretable, scale-consistent metric across
environments. To assess generalization, we report this metric on
held-out tasks not seen during training, and additionally perform
model rollouts to test how well the learned policy performs on states
that differ from those visited by the expert. We then examine
\emph{task resolution invariance}: the sensitivity of the operator's
predictions to the number and ordering of context points provided at
test time, including context set sizes not seen during training.

\begin{table}[h!]
\caption{Evaluation of the SetONet model across four OCP environments
and a reduced-data variant (P2P-Small), using a single expert
trajectory for demonstration and 32 trajectories for evaluation. The top section reports zero-shot predictions
from the pretrained and meta-trained operators with no gradient
updates. The lower sections report performance after 1 and 25
gradient steps of adaptation. Bold indicates the best result per
method and column. ``--'' indicates divergence. The pretrained operator achieves
low zero-shot error on P2P-Cost and Quadrotor, while
dynamics-varying and obstacle environments benefit from
adaptation, particularly via meta-trained initializations. All errors are reported as relative $L^2$ \cref{eq:rel_l2} }
\label{tab:adaptation_results}
\centering
\small
\begin{tabular}{l ccccc}
    \toprule
    Method & P2P-Cost & P2P-Small & P2P-Dyn. & Quadrotor & Obstacle \\
    \midrule
    \multicolumn{6}{l}{\textit{0 steps (zero-shot)}} \\
    Pretrained
      & \textbf{.048} & .101 & \textbf{.179} & \textbf{.063} & \textbf{.238} \\
    MAML
      & .583 & .920 & .802 & .549 & 1.00 \\
    SetONet-Meta
      & .075 & \textbf{.084} & .276 & .185 & .287 \\
    SetONet-Meta-Full
      & .118 & .130 & .195 & .148 & .395 \\
    \midrule
    \multicolumn{6}{l}{\textit{1 gradient step}} \\
    SetONet-FT
      & .080 & .090 & .238 & .069 & \textbf{.232} \\
    Last-Branch
      & \textbf{.045} & .095 & .180 & .061 & .238 \\
    Last-Both
      & .047 & .088 & .182 & \textbf{.059} & .237 \\
    MAML
      & .581 & .917 & .898 & .121 & 1.00 \\
    SetONet-Meta
      & .074 & \textbf{.083} & .255 & .170 & .287 \\
    SetONet-Meta-Full
      & .117 & .129 & \textbf{.099} & .066 & .311 \\
    \midrule
    \multicolumn{6}{l}{\textit{25 gradient steps}} \\
    SetONet-FT
      & .065 & .090 & .143 & .065 & \textbf{.237} \\
    Last-Branch
      & \textbf{.044} & \textbf{.075} & .177 & .059 & .231 \\
    Last-Both
      & .048 & .076 & .171 & \textbf{.057} & .234 \\
    MAML
      & .569 & .898 & -- & .118 & 1.00 \\
    SetONet-Meta
      & .069 & .077 & .094 & .071 & .285 \\
    SetONet-Meta-Full
      & .096 & .102 & \textbf{.081} & .073 & .306 \\
    \bottomrule
\end{tabular}
\end{table}

\paragraph{Fitting the operator.}
For each environment, we train a SetONet on a distribution of tasks
and evaluate on held-out tasks not seen during training. At test time,
the trained operator receives a context set from the new task and
predicts the corresponding feedback policy. We report results under
two evaluation settings: \emph{expert-states}, in which the model predicts
controls at the same states visited by the expert, and
\emph{model-rollout}, in which the predicted policy is executed from
the same initial condition without any expert feedback. The
model-rollout setting is the more demanding test, as it exposes the
model to compounding errors---a well-known challenge in imitation
learning often referred to as \emph{distributional
shift}~\citep{ross2011reduction}, where small prediction errors
accumulate and push the state trajectory away from the expert's
distribution.
 
To quantify expert-states accuracy, we compute the behavioral cloning
loss~\cref{eq:bc_loss} across all expert trajectories for each task
and report the relative $L^2$ error \cref{eq:rel_l2}. The zero-shot rows of
\cref{tab:adaptation_results} (0 steps, Pretrained) report these
errors for all four OCP environments and P2P-Cost-small. The pretrained operator achieves low relative error
on P2P-Cost and Quadrotor, where the task variation is
well-covered by the training distribution. P2P-Cost required more
demonstration trajectories per task to achieve comparable error, which we attribute to
the wide range of goal locations ($\state_g \in [-10,10]^2$):
distant goals produce larger control signals and correspondingly
noisier relative error. The dynamics-varying (P2P-Dynamics) and obstacle environments show higher pretrained error, reflecting the greater diversity of these task distributions; these are the settings where adaptation yields
the largest gains, as detailed in \cref{sec:adaptation}.
 
\begin{figure}[h!]
  \centering
  \includegraphics[width=0.65\linewidth]{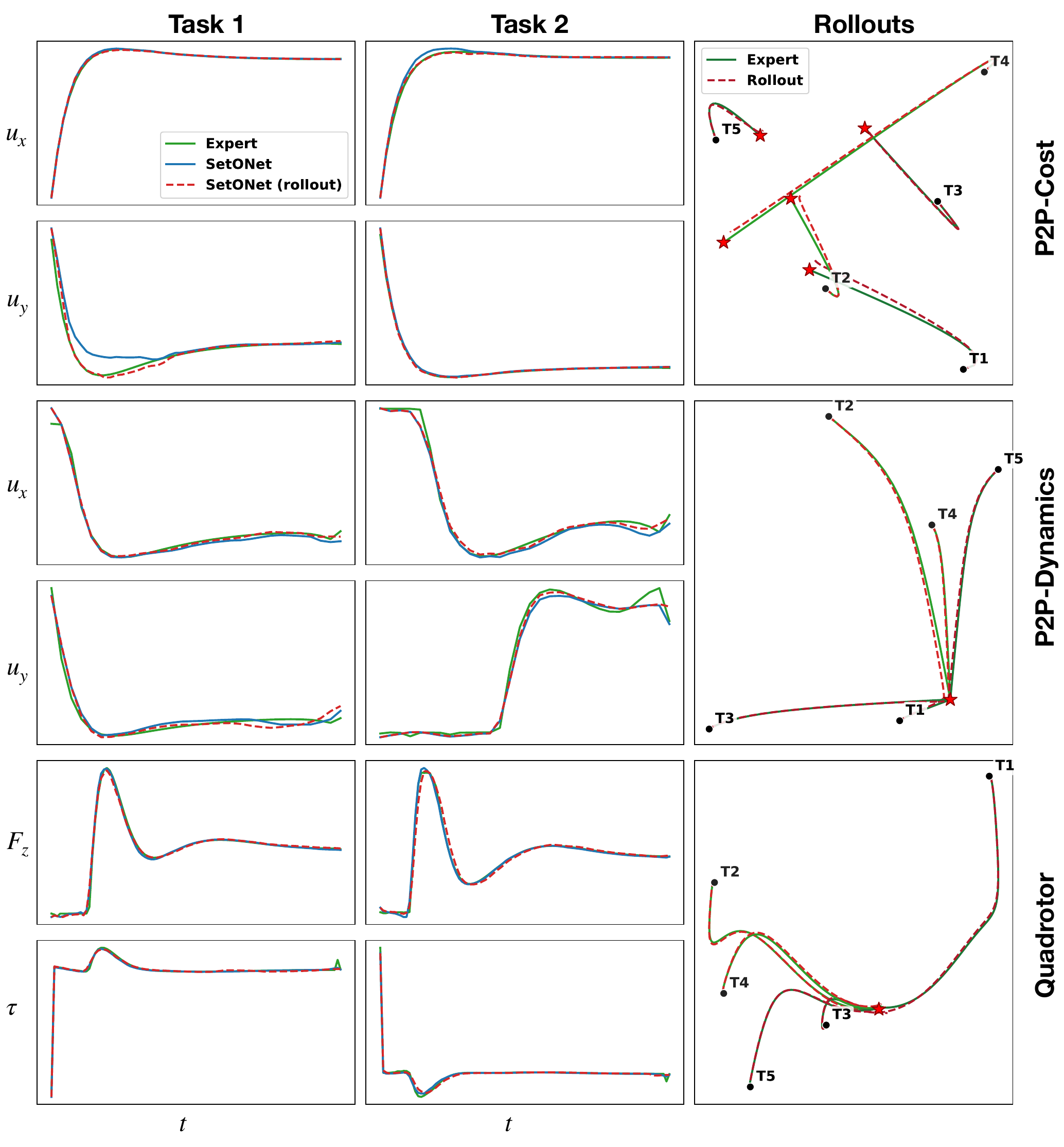}
  \caption{Operator fitting results across three environments.
    Each row group shows two control dimensions for a given environment,
    with columns displaying predictions on two representative tasks.
    We then show the corresponding state-space rollouts for the two tasks (T1, T2) along with 3 more (T3-T5).
    Solid green lines denote expert demonstrations, solid blue are
    SetONet predictions at the expert state locations, and dashed
    lines denote model-rollouts using the learned operator policy.}
  \label{fig:operator_fitting}
\end{figure}

\cref{fig:operator_fitting} provides a qualitative view of these
results across three environments, overlaying expert-states
predictions and model-rollout controls on the expert for
representative held-out tasks. Across all three environments, the
learned operator produces model-rollout trajectories that closely
follow the expert without any online correction. The Planar
Quadrotor presents the most challenging fitting problem due to its
higher-dimensional state space and coupled dynamics, yet the model
still captures the qualitative control profile and the rollouts
reach the correct targets despite minor transient deviations.

\paragraph{Task resolution invariance.}
A key property of the set-based context representation is that the
learned operator can be evaluated with context sets of varying
cardinality at test time, without retraining. This \emph{task
resolution invariance} arises because the branch network learns a
set-based representation of the task from the context set, rather than
relying on a fixed ordered discretization. The permutation-invariant
aggregation ensures that this representation is well defined for
variable-sized, unordered context sets.
This is particularly appealing in control settings, where the amount of expert data may vary across tasks (e.g., in the number of trajectories or rollout lengths), allowing the model to flexibly leverage available data at inference time without retraining
 
% increase font isze 
\begin{figure}
    \centering
    \includegraphics[width=0.7\linewidth]{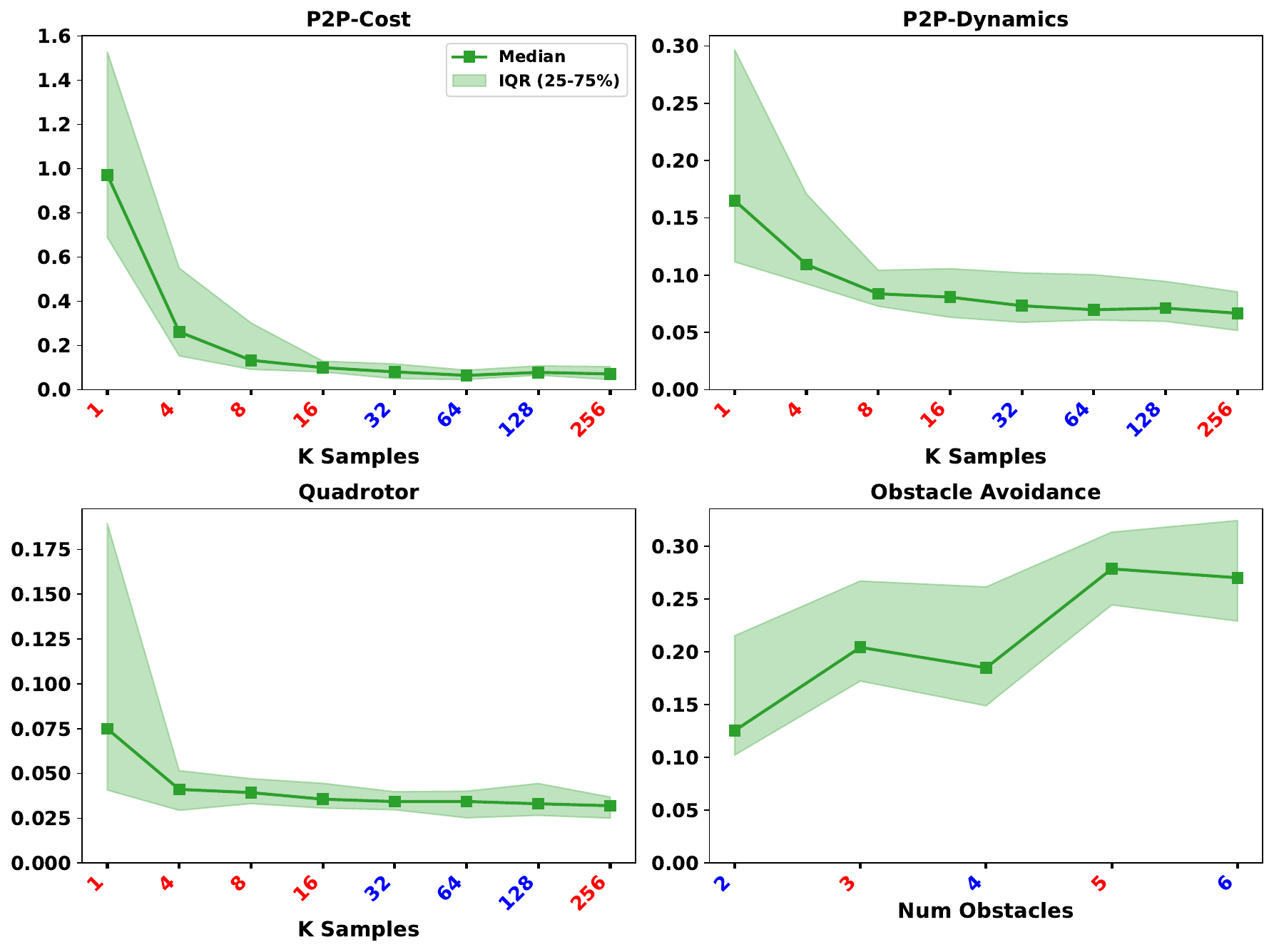}
     \caption{Task resolution invariance across all four control environments.
Lines show median relative $L^2$ error over held-out tasks; shaded
regions indicate the inter-quartile range. Blue tick marks denote
context sizes seen during training; red tick marks denote sizes not
seen during training. The $x$-axis for the first three environments
is the number of context samples; for Obstacle Avoidance it is the
number of obstacles. The model was trained on obstacle configurations
with 2, 4, and 6 obstacles.}
    \label{fig:task_resolution}
\end{figure}

\cref{fig:task_resolution} evaluates this property across all four
environments. In each panel, the $x$-axis varies the number of
context samples provided to the branch encoder at test time,
including sizes both seen and not seen during training. Lines show
the median relative $L^2$ error over held-out tasks and shaded
regions indicate the interquartile range (IQR). We report median
and IQR rather than mean and standard deviation because the error
distribution across tasks is right-skewed: a small number of
particularly challenging configurations can produce large errors
that disproportionately inflate the mean.

For P2P-Cost, P2P-Dynamics, and the Planar Quadrotor, all three
environments exhibit a consistent pattern: error is noticeably
higher and more variable with very few context points (1--4), but
decreases steadily and stabilizes by around 16--32 samples. Beyond
this point, performance remains steady across the full range,
including at context sizes not seen during training. This indicates
that the branch encoder requires only a modest number of samples to
extract reliable task information, and does not overfit to a
particular context cardinality. In the dynamics-varying environments
(P2P-Dynamics, Quadrotor), the initial decrease is expected: a
single observation of
$\state_{t+1} = \dynam(\state_t, \ctrl_t; \dynamparam)$ provides
limited information about the dynamics parameters, but a handful of
samples suffices to approximately identify them. P2P-Cost exhibits a
similar pattern, with error dropping sharply from 1 to 8 context
points before leveling off.

The Obstacle Avoidance environment (\cref{fig:task_resolution},
bottom right) presents a qualitatively different picture. Unlike
the other three environments, the median error increases with the
number of obstacles. We attribute this to the inherent growth in
task complexity: configurations with more obstacles create tighter
passages and more constrained feasible trajectories, making the
mapping from obstacle configuration to collision-free policy harder
to approximate. The widening IQR at higher obstacle counts further
suggests that some configurations are significantly harder than
others, likely those with narrow corridors or near-degenerate
passages. Notably, the model maintains reasonable accuracy at
obstacle counts not seen during training (3 and 5 obstacles,
red ticks), demonstrating that the set-based representation
interpolates smoothly between the training configurations
($n_{\text{obs}} \in \{2, 4, 6\}$). 
 
This resolution invariance is a practical advantage: at deployment,
the operator can produce reasonable policies from whatever context
data is available, whether more or fewer samples than were used
during training.
 
\begin{figure}[t]
    \centering
    \includegraphics[width=1\linewidth]{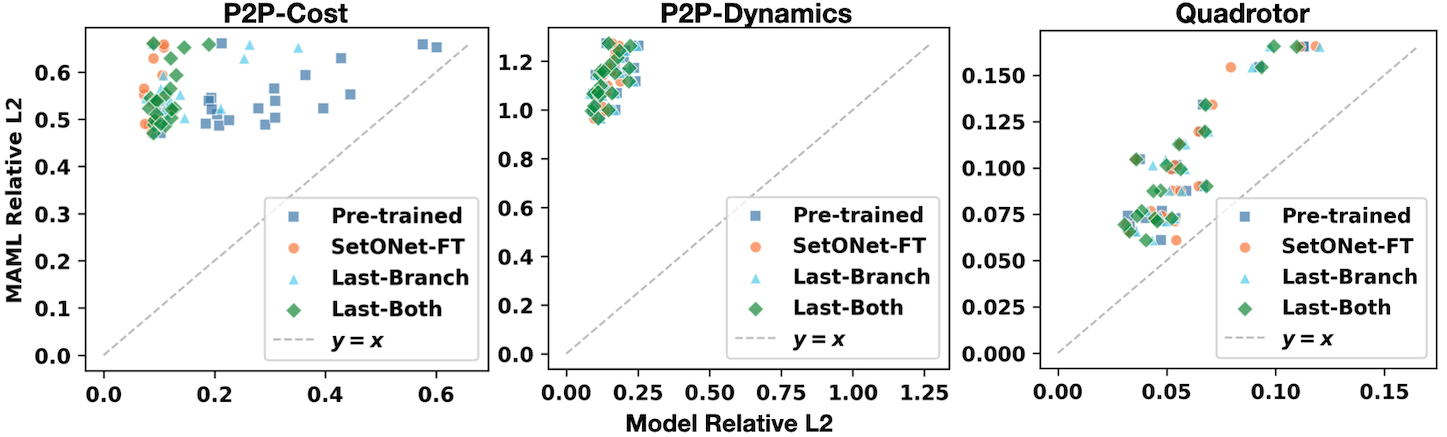}
    \caption{Per-task comparison of MAML against four SetONet-based
methods across three OCP environments after 25 gradient steps
using 10 expert trajectories for adapting and evaluation. Each point represents a
single held-out task, with the $x$-axis showing the method's
relative $L^2$ error and the $y$-axis showing MAML's error on
the same task. Points above the diagonal ($y = x$) indicate
tasks where MAML performs worse. MAML is consistently dominated
across nearly all tasks in P2P-Cost and P2P-Dynamics, while in
the Quadrotor environment the gap narrows, with some tasks
falling near the diagonal.}
    \label{fig:transfer_summary}
\end{figure}

\subsection{Task-Specific Adaptation}
\label{sec:adaptation}
We now evaluate the adaptation strategies introduced in \cref{sec:transfer} on the held-out tasks not seen during training. This section focuses on fine-tuning using expert demonstrations (compared against a MAML baseline) and adaptation using cost feedback. Meta-trained operators are evaluated separately in \cref{sec:exp:meta_op}. A central objective of this evaluation is to provide a systematic comparison between neural operator adaptation and meta-learning approaches popular in control and reinforcement learning, such as MAML. Through these comparisons, we demonstrate that efficient fine-tuning of neural operators is competitive, and often superior to MAML, without incurring additional meta-training cost.
 
\textbf{Adaptation with expert demonstrations.}
\cref{tab:adaptation_results} reports the mean relative $L^2$
error across five OCP environments after 0, 1, and 25 gradient
steps of adaptation using a single expert demonstration (standard deviations across seeds were negligible and are omitted for clarity). Among the non-meta-trained methods,
Last-Branch and Last-Both, which update only the last layer of
the branch network or the last layer of both networks,
respectively, consistently perform well: a single gradient step
is often sufficient to match or improve upon the pretrained
operator, and 25 steps yield further gains on the
dynamics-varying environments (P2P-Dynamics, Quadrotor).
SetONet-FT, which updates the full network, shows the
largest improvements on P2P-Dynamics, reducing error from .179
to .143 after 25 steps, but provides little benefit on P2P-Cost
and Obstacle where the pretrained model is already accurate.
 
MAML, by contrast, produces high error across all environments and
step counts, often failing to improve over its poor zero-shot
initialization. \cref{fig:transfer_summary} provides a per-task
view of this gap: each point compares MAML's error against one of
the four SetONet-based methods on the same held-out task. In
P2P-Cost and P2P-Dynamics, nearly all points lie above the
diagonal, indicating that MAML is dominated on virtually every
individual task. In the Quadrotor environment the margin narrows,
with some tasks falling near the diagonal, though the SetONet
methods still hold an overall advantage. 

\textbf{Adaptation with cost feedback.}
In some settings, expert
demonstrations for the target task may be unavailable, the expert
may have been trained on a cost function that we would like to
modify, or only suboptimal policies may be accessible. Cost-based
fine-tuning addresses these cases by adapting the operator directly
on a downstream cost function, assuming knowledge of both the cost and
environment dynamics. The operator is adapted by
differentiating through model-rollouts, as described in
\cref{sec:transfer}.

\begin{figure}[t]
\centering
\begin{subfigure}[t]{\textwidth}
    \centering
    \includegraphics[width=\textwidth]{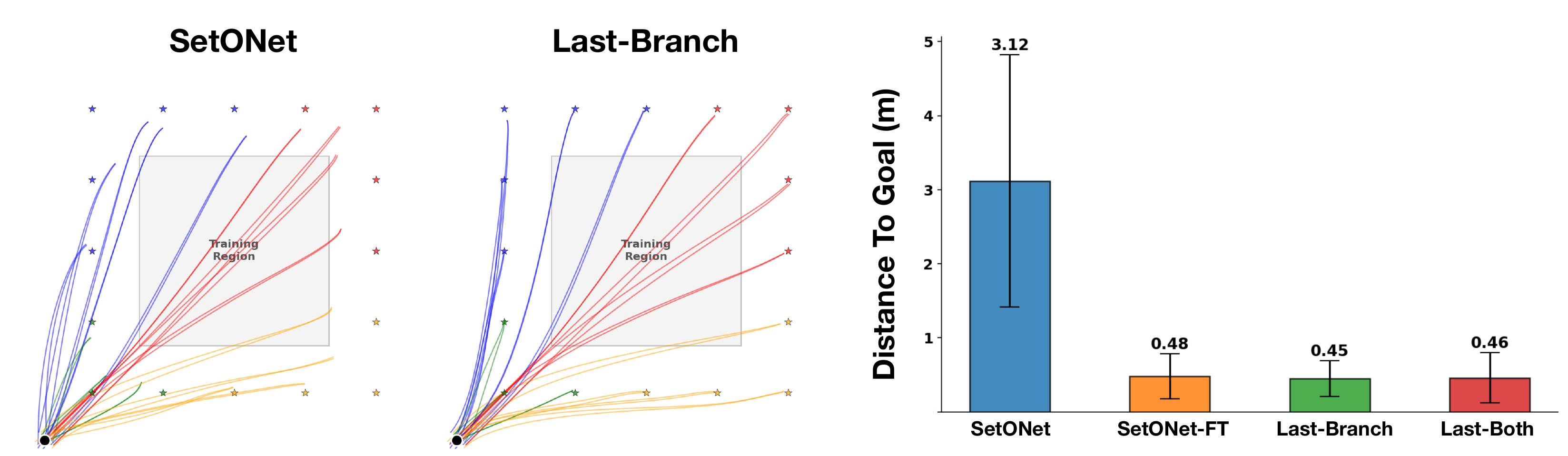}
    \caption{P2P-Cost (out-of-distribution, cost-based adaptation)}
    \label{fig:cost_adapt_p2p}
\end{subfigure}
\vspace{0.3cm}
\begin{subfigure}[t]{\textwidth}
    \centering
    \includegraphics[width=\textwidth]{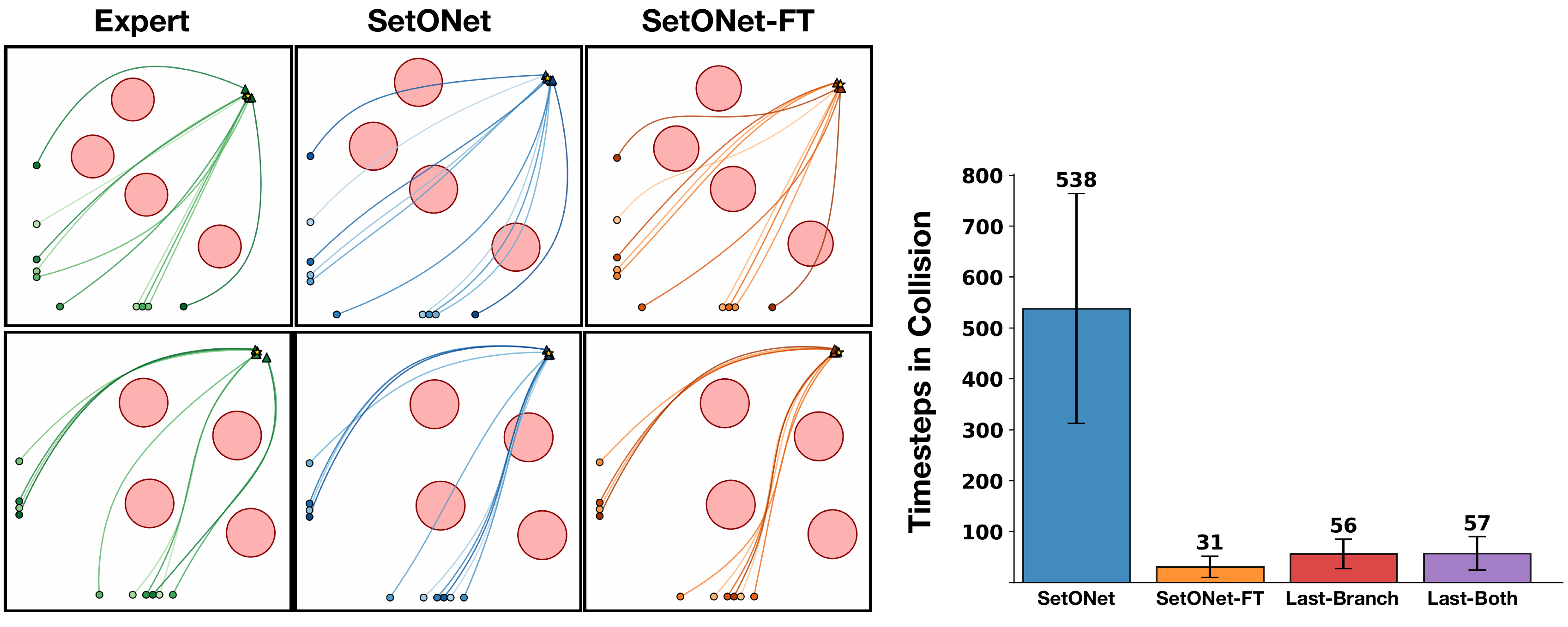}
    \caption{Obstacle Avoidance (held-out tasks, cost-based adaptation)}
    \label{fig:cost_adapt_obstacle}
\end{subfigure}
\caption{Cost-based fine-tuning across two environments. \textbf{(a)}~Out-of-distribution
P2P-Cost tasks with both goal locations and start position outside the training region (gray). The distance is $L^2$ in meters. The pretrained SetONet fails to reach distant goals (mean terminal distance of 3.12. After adaptation, all methods recover accurate trajectories, with Last-Branch (0.45) slightly
outperforming SetONet-FT (0.48), indicating that the pretrained basis functions (trunk network) generalize
well and only the coefficients require updating. \textbf{(b)}~Held-out obstacle avoidance
tasks. All cost-based
fine-tuning strategies reduce collision time well below the pretrained model}
\label{fig:cost_adaptation}
\end{figure}

\cref{fig:cost_adapt_p2p} illustrates this on out-of-distribution
P2P-Cost tasks. The surrogate is the same LQR objective used
during training (\cref{eq:lqr_cost}), differentiated end-to-end
through the known linear dynamics. After fine-tuning, all three
strategies (full network, branch-only, and last-layer ) converge
to costs comparable to the expert, demonstrating that the operator
can be adapted purely from the task objective without any expert
data.

\cref{fig:cost_adapt_obstacle} applies cost-based fine-tuning to
held-out obstacle avoidance tasks. Here the surrogate cost
combines a soft collision penalty,
$w_{\mathrm{coll}}\sum_{i}\exp\!\bigl(-\alpha(d_{i,t}-r_i-m)\bigr)$,
where $d_{i,t}$ is the distance from the agent to obstacle~$i$,
$r_i$ is the obstacle radius, and $m=0.2$ is a safety margin,
with a control regularization term $w_{\mathrm{ctrl}}\lVert
u_t\rVert^2$ and a terminal goal-reaching penalty, evaluated
along a differentiable rollout through double-integrator
dynamics ($w_{\mathrm{coll}}=10$, $w_{\mathrm{ctrl}}=\Delta t$,
$\alpha=15$). The pretrained SetONet spent an average of
${\sim}3.5\times$ the number of collision timesteps as the expert.
All three fine-tuning strategies reduce collision time well below
the expert while successfully reaching the goal on every task.
Because the expert solver enforces collision avoidance through
hard constraints rather than directly minimizing collision time,
the fine-tuned models can achieve fewer collision timesteps by
explicitly penalizing proximity to obstacles in the surrogate
cost.
Together, these two experiments demonstrate that cost-based
adaptation can refine the operator directly from the task
objective, without requiring any expert demonstrations.

\subsection{Meta-trained operator}\label{sec:exp:meta_op}
The fine-tuning results above rely on a model that was pretrained
with the standard behavioral cloning objective. We now evaluate
whether the meta-training procedure of \cref{sec:meta_training},
which explicitly optimizes for rapid post-adaptation performance,
can improve adaptation efficiency.
 
Returning to \cref{tab:adaptation_results}, the zero-shot rows
reveal a clear difference between the two meta-trained variants.
SetONet-Meta achieves the lowest zero-shot error on P2P-Small
(.084) and competitive error on P2P-Cost (.075), outperforming
SetONet-Meta-Full in both cases. On P2P-Dynamics, however,
SetONet-Meta-Full produces the best zero-shot result (.195),
suggesting that the ability to adapt basis functions, not just
coefficients, matters when the task variation is more complex.
After 25 gradient steps, SetONet-Meta-Full achieves the lowest
error on P2P-Dynamics (.081), while SetONet-Meta remains
competitive on P2P-Small (.077). Both meta-trained variants
underperform Last-Branch and Last-Both on P2P-Cost and Quadrotor,
where the pretrained operator is already accurate and the
additional meta-training offers little benefit.
 
\begin{figure}
    \centering
    \includegraphics[width=1\linewidth]{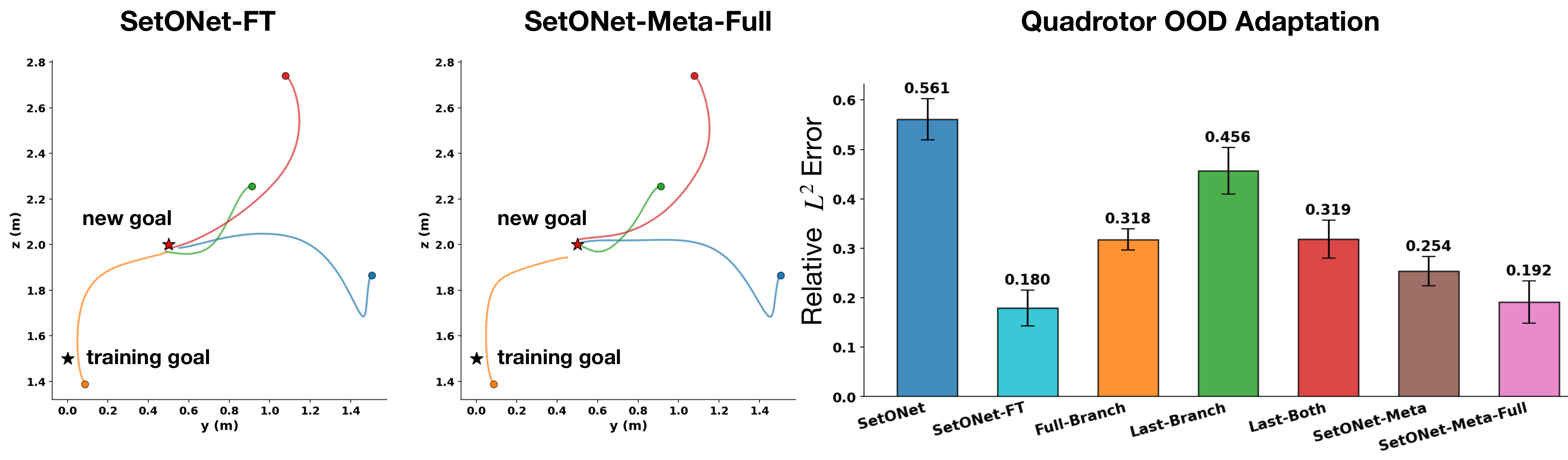}
    \caption{Out-of-distribution fine tuning on a Quadrotor task. Left two panels show multiple representative trajectories for SetONet-FT and SetONet-Meta-Full adapting to a shifted goal location. The bar plot
    compares relative $L^2$ error across all fine-tuning and meta-learning methods.}
    \label{fig:quadrotor_ood}
\end{figure}

We also evaluated meta-trained adaptation on
an out-of-distribution (OOD) Quadrotor task in which the goal location
is shifted outside the training region.
\cref{fig:quadrotor_ood} compares all adaptation methods on this
task. The pretrained SetONet as expected exhibits the largest error. 
SetONet-FT and SetONet-Meta-Full achieve the lowest errors, demonstrating the importance 
of trunk adaptation on an OOD task. We see that the meta-trained initialization enables effective adaptation even when the target task lies outside the training distribution.
This is a setting where the pretrained basis functions do not
generalize well, and adapting both coefficients and basis functions
provides a clear advantage. 

\begin{figure}
    \centering
    \includegraphics[width=1.0\linewidth]{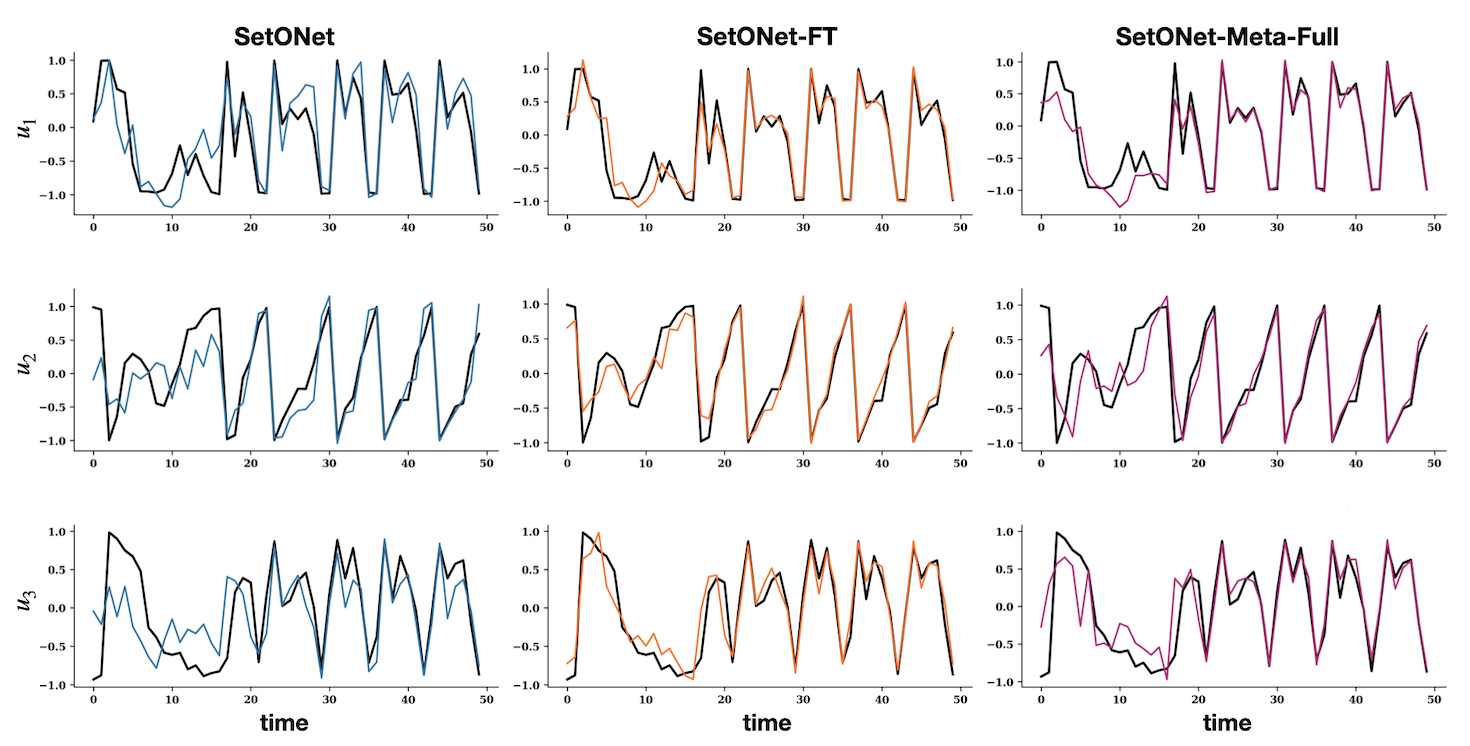}
    \caption{Control predictions on a held-out HalfCheetah-v3
task, showing the first three control dimensions
($u_1, u_2, u_3$, rows) for three methods (columns) over time.
Each panel compares the method's prediction (colored) against
the expert (black) after 25 gradient steps using 10 expert
trajectory demonstrations of 100 timesteps each. SetONet-FT most closely tracks the expert across all dimensions.}
    \label{fig:cheetah_control}
\end{figure}

\textbf{Meta-training on HalfCheetah-v3.}
We next evaluate the meta-trained operators on
HalfCheetah-v3, where expert policies are trained via
reinforcement learning (SAC). \cref{fig:cheetah_control} shows the first three
control dimensions for representative held-out HalfCheetah-v3
configurations. Unlike the smooth, low-frequency control signals
of the OCP environments, the expert policies here produce
high-frequency, multi-dimensional outputs that the operator must
approximate from noisy, suboptimal demonstrations, making
accurate few-shot adaptation considerably more demanding.
 
\begin{figure}
    \centering
    \includegraphics[width=0.8\linewidth]{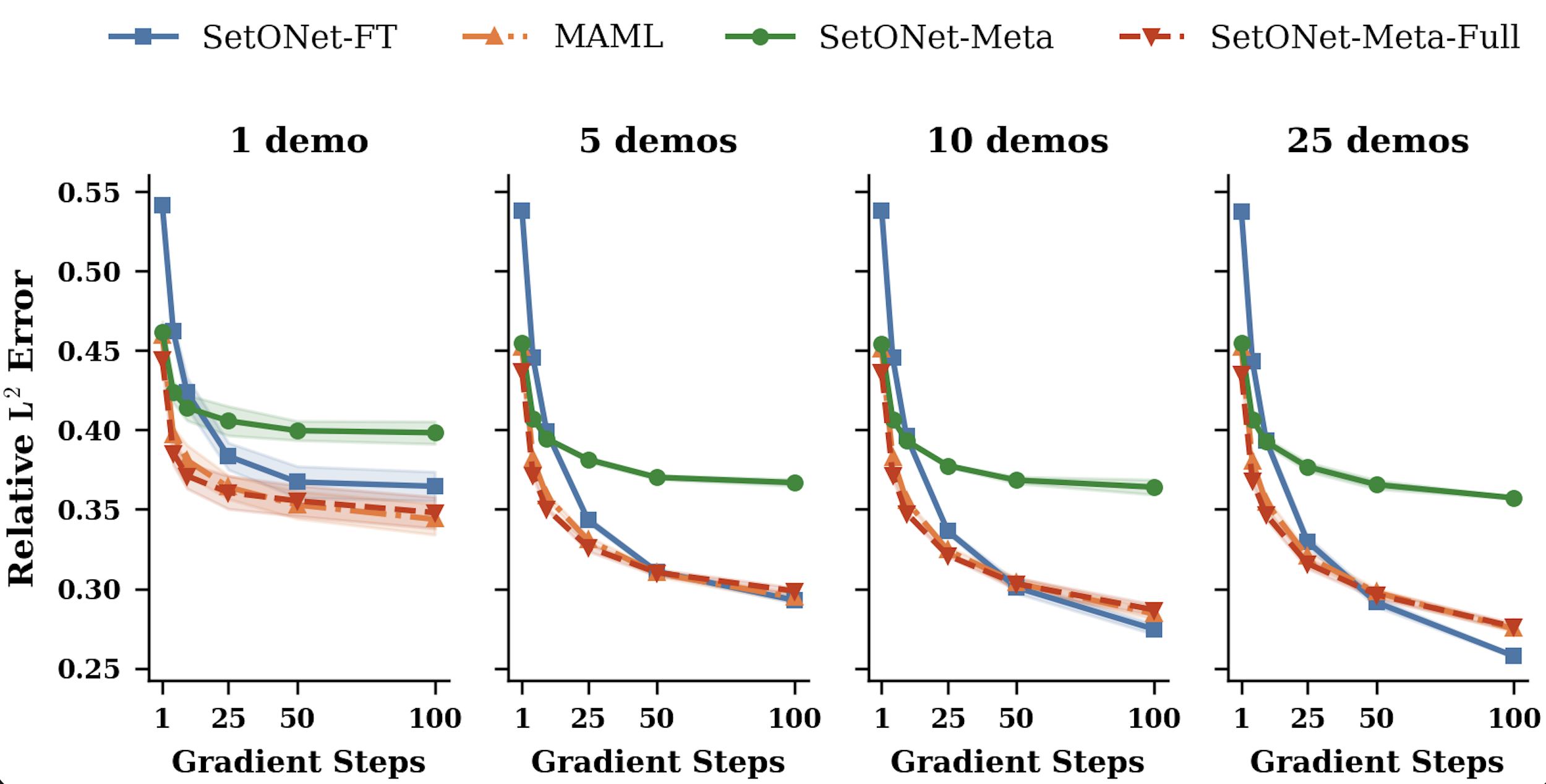}
    \caption{Adaptation performance on held-out HalfCheetah-v3
tasks as a function of the number of expert demonstrations
(panels) and gradient steps (horizontal axis). Lines show mean
relative $L^2$ error over 5 seeds; shaded regions indicate $\pm 1$
standard deviation. With few gradient steps (1--5), the meta-trained variants outperform SetONet-FT, particularly
in the low-data regime (1 and 5 demos). As the adaptation budget
increases, MAML and SetONet-Meta-Full improve steadily while SetONet-FT has the largest decrease in error as the number of available samples is increased. }
    \label{fig:halfcheetah_grid}
\end{figure}

\cref{fig:halfcheetah_grid} examines these trade-offs in greater
depth, comparing four adaptation strategies across varying numbers
of expert demonstrations (panels) and gradient steps (horizontal
axis). Several patterns emerge. First, MAML and SetONet-Meta-Full
consistently outperform SetONet-FT in the low-data regime (1 and 5
demonstrations), achieving lower error across all gradient step
budgets. Since both MAML and SetONet-FT perform full-network
adaptation, this advantage is attributable to the meta-trained
initialization rather than the scope of parameters being updated:
the bi-level training objective produces a starting point from
which a single gradient step already yields a strong task-specific
model. Second, SetONet-FT benefits most from additional
demonstrations and gradient steps, closing the gap with the
meta-trained methods at 10 demonstrations and overtaking them at
25 demonstrations with 50 or more gradient steps. This is
expected---with sufficient data, the quality of the initialization
matters less, and SetONet-FT is free to converge to any solution
without constraints imposed by the inner-loop architecture chosen
at meta-training time. Third, SetONet-Meta plateaus early,
confirming that restricting the inner loop to the branch network
alone limits adaptation capacity in this higher-dimensional
environment.
 
These results highlight a practical trade-off: when the adaptation
budget is limited to a small support set and few gradient steps,
the meta-trained initialization provides a significant advantage
over standard pretraining. As more data and compute become
available, standard fine-tuning catches up and eventually achieves
the lowest error. The choice between strategies thus depends on the
deployment setting: meta-training is preferable for rapid,
low-resource adaptation, while standard fine-tuning is preferable
when a larger adaptation budget is available. Practical choices can be made problem-specific, in cases where rapid task switching is required during online deployment, meta-training of the neural operator controller can provide a significant boost in efficiency.

\section{Conclusion}
 
We have presented a general framework for multi-task optimal control based
on neural operators, demonstrating that the operator learning
perspective offers a principled and practical approach to learning
control policies across families of related tasks. By modeling the
mapping from task-defining functions to optimal feedback policies,
neural operators naturally capture the shared structure underlying
parametric control problems while remaining flexible enough to
accommodate diverse task variations.

Here we highlight a number of key results from this work. First, we showed through various experiments how the SetONet model can closely approximate the solution operators defined in \eqref{eq:soln_operator}. We demonstrated different forms of generalization on held-out tasks, including model rollouts that closely matched the expert and task resolution invariance, a property that is particularly valuable when the amount of available task information varies at deployment. Second,
the branch--trunk decomposition enables effective task-specific
adaptation, and we presented a number of adaptive model choices that target different parts of the network. We discussed the trade-offs of each, which depend on the available training data and its relationship to the downstream task. Third, we introduced
two meta-training variants, SetONet-Meta and
SetONet-Meta-Full, which explicitly optimize the initialization for
rapid few-shot adaptation. SetONet-Meta provides data-efficient updates by
restricting adaptation to the branch coefficients, while
SetONet-Meta-Full adapts both coefficients and basis functions,
yielding the largest gains on out-of-distribution tasks such as
P2P-Dynamics and HalfCheetah-v3. Both consistently outperform
MAML, which struggles to adapt in most environments. Finally, cost-based
adaptation demonstrated that the operator can be refined directly from the task objective without any expert demonstrations, using only the cost function and the dynamics model. 
 
Cost-based adaptation opens several directions for further exploration. For example, online data collection strategies such as
DAgger~\citep{ross2011reduction} could iteratively refine the
operator by querying the expert only in states actually visited under
the learned policy, improving closed-loop performance while
limiting the total number of expert solves. More broadly, although we
explored several strategies for adaptation and fine-tuning, many
remain unexplored, including methods designed specifically for
out-of-distribution settings~\citep{kumar2022fine}. In this work, we
restricted attention to settings in which a single input function
varies across tasks (either cost or dynamics). Extending the operator
learning framework to problems where tasks are defined by
simultaneous variation in multiple function spaces is an open and
practically relevant direction: for instance, in the point-to-point
environments, one could vary both the cost function and the dynamics
simultaneously. Recent work on meta-learning with neural operators ~\citep{wang2024meta} takes a \textit{multi-operator} approach to this problem, but focuses on a single application domain and does not provide theoretical analysis, leaving the general setting largely unexplored. 
 
\bibliography{main}
\bibliographystyle{tmlr}

\end{document}

%% file: math_commands.tex
\usepackage{amsmath,amsfonts,bm}

\def\eqref#1{equation~\ref{#1}}
\def\1{\bm{1}}

\DeclareMathAlphabet{\mathsfit}{\encodingdefault}{\sfdefault}{m}{sl}
\SetMathAlphabet{\mathsfit}{bold}{\encodingdefault}{\sfdefault}{bx}{n}